\documentclass[11pt]{article}

\usepackage[final]{acl}

\usepackage{times}
\usepackage{latexsym}
\usepackage{amsmath}
\usepackage{booktabs}
\usepackage{tabularx}
\usepackage{multirow} 
\usepackage{subcaption}
\usepackage{array} 
\usepackage[T1]{fontenc}

\usepackage[utf8]{inputenc}

\usepackage{microtype}

\usepackage{inconsolata}

\usepackage{graphicx}
\usepackage{multirow}
\usepackage{makecell}
\usepackage[table]{xcolor}
\usepackage{pifont}
\newcommand{\cmark}{\textcolor{green!40!black}{\ding{51}}}
\newcommand{\xmark}{\textcolor{red!70!black}{\ding{55}}}

\title{\textsc{UniKIE-Bench}: Benchmarking Large Multimodal Models for Key Information Extraction in Visual Documents}

\def\method{\textsc{UniKIE-Bench}}

\author{Yifan Ji$^{1}$\thanks{ \ \ indicates equal contribution.}, Zhipeng Xu$^{1}$\footnotemark[1], Zhenghao Liu$^{1}$\thanks{ \ \ indicates corresponding author.}, Zulong Chen$^{3}$, Qian Zhang$^{3}$, \\ \textbf{Zhibo Yang$^{3}$, Junyang Lin$^{3}$, Yu Gu$^{1}$, Ge Yu$^{1}$ and Maosong Sun$^{2}$}\\
$^1$School of Computer Science and Engineering, Northeastern University, Shenyang, China \\
$^2$Department of Computer Science and Technology, Tsinghua University, Beijing, China \\
$^3$Alibaba Group, Hangzhou, China\\
}

\begin{document}
\maketitle
\begin{abstract}
Key Information Extraction (KIE) from real-world documents remains challenging due to substantial variations in layout structures, visual quality, and task-specific information requirements. Recent Large Multimodal Models (LMMs) have shown promising potential for performing end-to-end KIE directly from document images. To enable a comprehensive and systematic evaluation across realistic and diverse application scenarios, we introduce \method{}, a unified benchmark designed to rigorously evaluate the KIE capabilities of LMMs. \method{} consists of two complementary tracks: a constrained-category KIE track with scenario-predefined schemas that reflect practical application needs, and an open-category KIE track that extracts any key information that is explicitly present in the document.
Experiments on 15 state-of-the-art LMMs reveal substantial performance degradation under diverse schema definitions, long-tail key fields, and complex layouts, along with pronounced performance disparities across different document types and scenarios. These findings underscore persistent challenges in grounding accuracy and layout-aware reasoning for LMM-based KIE. All codes and datasets are available at \url{https://github.com/NEUIR/UNIKIE-BENCH}.
\end{abstract}

\section{Introduction}

Key Information Extraction (KIE) aims to identify and extract critical fields from visually rich and semantically diverse documents~\cite{tang1994document,cesarini2002informys,mao2003document}. It serves as a fundamental component for downstream applications such as automated accounting, financial auditing, and enterprise knowledge management~\cite{nasar2018information,oral2020information,skalicky2022business}.
Despite extensive progress, KIE in real-world settings remains highly challenging due to substantial variations in document layouts, capture conditions, linguistic patterns, and field definitions~\cite{gbada2025deep}. 
Addressing these challenges requires fine-grained multimodal understanding capabilities that tightly couple visual structure with textual semantics~\cite{zhang2023reading,ding2024deep,rombach2025deep}.

Recent advances in Large Multimodal Models (LMMs) have renewed interest in KIE tasks~\cite{bai2025qwen2,comanici2025gemini,vafaie2025end}.
Unlike traditional approaches that rely on explicit and modular pipelines for text, layout, and visual processing, LMMs extract information from document images in an end-to-end manner. This design mitigates error propagation across modules and enables strong generalization across a wide range of visual tasks~\cite{liu2024ocrbench,bai2025longbench,yang2025cc}.
Despite these advances, existing benchmarks remain insufficient for comprehensively evaluating LMMs' KIE capabilities under realistic document settings~\cite{ebrahimi2022test, he2023good, wei2025p2net}.
In particular, most KIE benchmarks are designed for specific application scenarios and rely on heterogeneous annotation schemas, task formulations, or evaluation metrics, making it difficult to support consistent, end-to-end evaluation across diverse scenarios~\cite{laatiri2023information, kuang2023visual}.
Furthermore, the predefined fields in these datasets are often tightly coupled to scenario-specific requirements, limiting their ability to systematically cover the diverse types of key information encountered in real-world documents~\cite{laatiri2023information, kuang2023visual, shi2023exploring}.

\begin{figure*}
    \centering
    \includegraphics[width=\linewidth]{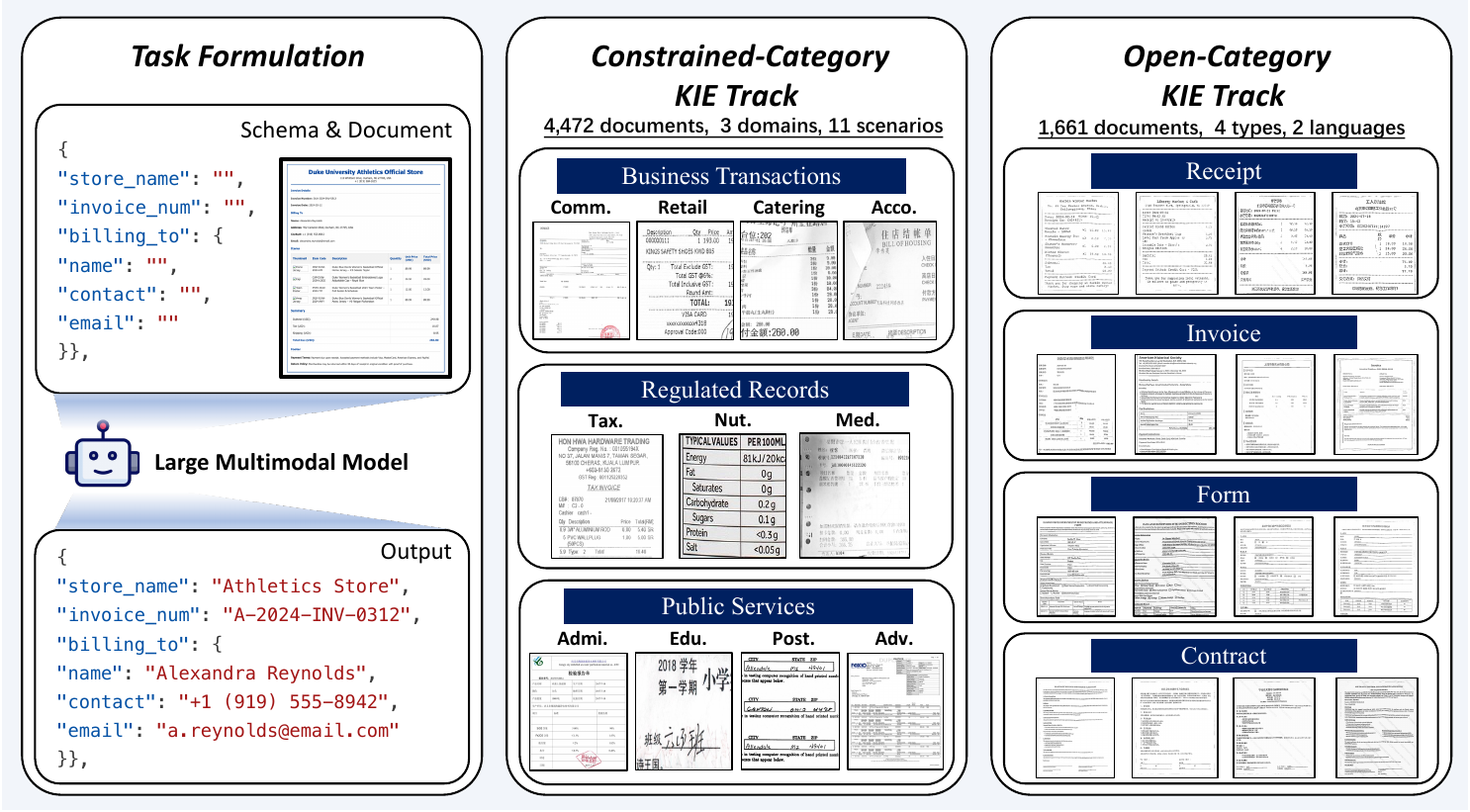}
    \caption{Overview of our \method{}. \method{} is built upon a schema-guided KIE formulation to enable end-to-end evaluation of LMMs. It comprises two complementary evaluation tracks: the constrained-category KIE track and the open-category KIE track.}
\label{fig:main_figure}
\end{figure*}

To address these limitations, we introduce \method{}, a comprehensive benchmark for the systematic evaluation of KIE capabilities in LMMs. As shown in Figure~\ref{fig:main_figure}, \method{} is built upon a unified, schema-guided structured prediction formulation, which enables a consistent evaluation paradigm across diverse document types and annotation standards. Specifically, \method{} comprises two tracks: a constrained-category KIE track targeting specific application scenarios, and an open-category KIE track designed for general-purpose key information extraction.
The constrained-category KIE track incorporates task-specific information requirements derived from a wide range of real-world applications. These requirements are organized by their application scenario, enabling fine-grained, scenario-aware evaluation under well-defined schemas. In contrast, the open-category KIE track requires models to extract all key information explicitly present in each document, evaluating the general-purpose extraction capability of LMMs. 
Together, these two tracks enable \method{} to capture both targeted, application-driven performance and holistic extraction ability across heterogeneous documents.

Our experiments on 15 LMMs demonstrate that state-of-the-art LMMs still suffer from substantial limitations in KIE. In particular, performance degrades more severely when confronted with diverse schemas, long-tail fields, and complex document layouts, and this degradation is further exacerbated by realistic visual artifacts.
We further observe pronounced performance disparities across document types, application scenarios, and field categories, revealing persistent challenges in schema grounding, layout-aware reasoning, and robust information extraction under different visual conditions. These findings underscore the necessity of a unified and comprehensive benchmark that captures the structural and visual complexities of real-world documents. We hope that \method{} will facilitate more rigorous evaluation of LMM-based KIE systems and provide a solid foundation for assessing the generalizability and reliability of LMMs in document understanding.
\section{Related Work}
Key Information Extraction (KIE) is a fundamental task in document understanding, aiming to extract structured fields from visually rich documents~\cite{aiello2002document,gao2011structure,kim2022ocr}. 
Earlier research typically depends on Optical Character Recognition (OCR) to recognize textual content in the document images and employs graph neural networks or sequence labeling models to detect semantic relationships within the text~\cite{palm2017cloudscan,yu2021pick,wang2021vies,hui2025seg}. 
While effective, these methods rely heavily on accurate OCR results, leading recognition errors to cascade into field extraction~\cite{kim2022ocr,cheng2022trie++,dhouib2023docparser}. Moreover, OCR text fails to adequately preserve the original layout features, which may weaken or omit layout cues beneficial for extraction~\cite{lee2022formnet,liu2024see}. While some works explore encoding layout features into language models to enable layout-aware language modeling~\cite{xu2020layoutlm,wang2024docllm,huang2022layoutlmv3}, their performance remains tightly dependent on OCR accuracy~\cite{zhang2025ocr,barboule2025survey}.

More recent efforts have explored using Large Multimodal Models (LMMs) to capture key information from document images~\cite{ke2025large,wang2025document}.
Previous work adapts LMMs to text-rich documents by tailoring their visual encoders to effectively model dense text and complex layouts~\cite{liu2024hrvda,yu2024texthawk,zhang2025dockylin}. 
These methods typically employ filtering modules to remove content-irrelevant visual tokens~\cite{liu2024hrvda,zhang2025dockylin,liu2024textmonkey} or aggregate multi-scale visual features~\cite{park2024hierarchical,hu2025mplug} to enhance the visual text perception of LMMs~\cite{ding2024deep}. 
Recent research focuses on optimizing LMMs with document-oriented training objectives to enhance their document understanding capability~\cite{ye2023ureader,nacson2025docvlm}. 
MiniCPM-V adopts an OCR-oriented pretraining strategy that corrupts textual regions in document images and trains the model to reconstruct the original text~\cite{yu2025minicpm}. 
Kosmos-2.5 and Qwen2.5-VL further convert document pages into Markdown or HTML codes, offering a unified alignment target for text recognition and layout perception~\cite{lv2023kosmos,bai2025qwen2}.  

Despite recent progress in LMMs, evaluating their capabilities in KIE tasks remains challenging~\cite{shi2023exploring}. Early KIE benchmarks focus on a specific document type, such as receipts, invoices, and orders~\cite{huang2019icdar2019,vsimsa2023docile,abdallah2024survey}.
To enable broader evaluation, several recent document understanding benchmarks incorporate multiple KIE datasets into their evaluation suites~\cite{ding2024deep,ke2025large}. 
OCRBench introduces a dedicated KIE track that evaluates receipt and bill understanding by requiring LMMs to answer queries for predefined fields, thereby formulating KIE as a constrained question answering task over document images~\cite{liu2024ocrbench,fu2024ocrbench}. 
CC-OCR further expands the evaluation to business documents and requires LMMs to generate result dict conditioned on predefined field schemas, rather than answering independent queries field by field~\cite{yang2025cc}. Despite their contributions, these benchmarks primarily focus on homogeneous document types and narrow scenarios, limiting their ability to evaluate KIE performance on real-world documents comprehensively.

\section{\method{}: A Comprehensive Layout-Variant KIE Benchmark}
In this section, we formalize the Key Information Extraction (KIE) task and introduce the evaluation tracks in Sec.~\ref{bench:task}. We then present detailed statistics of the proposed benchmark in Sec.~\ref{bench:analysis}. The data curation process is described in Sec.~\ref{bench:data}. Finally, we compare our benchmark with existing KIE benchmarks in Sec.~\ref{bench:comparsion}.

\subsection{Task Formulation of KIE}
\label{bench:task}
To address the KIE task, existing KIE benchmarks~\cite{liu2024ocrbench,fu2024ocrbench} typically adopt a question-answering style formulation.
Specifically, given a document image $x$ and a set of target fields $\mathcal{F}$, the inquiry for each field
$f \in \mathcal{F}$ is verbalized as a field-specific query $q(f)$, and the model $\mathcal{M}$ generates a response $y_f$ to predict the value of each field independently:
\begin{equation}
\small
y_f= \mathcal{M} (x, q(f)),
\
f \in \mathcal{F}.
\end{equation}
The final prediction $\mathbf{y}^{\text{QA}} $ is obtained by aggregating all field-level outputs:
\begin{equation}
\small
\mathbf{y}^{\text{QA}} =
\{ f: y_f\}_{f \in \mathcal{F}}.
\end{equation}
While simple and flexible, this formulation requires multiple independent inferences for KIE and fails to capture the structural relationships among fields. 

In contrast, \method{} formulates KIE as a schema-guided structured prediction task that performs extraction in one inference step.
Each instance is represented by a schema $s = (\mathcal{F}, \mathcal{R})$, where  $\mathcal{F}$ denotes the set of target fields, and $\mathcal{R}$ specifies the relations among them. The model is required to generate a structured output based on the document image $x$ and the schema $s$:
\begin{equation}
\small
\mathbf{y}^{\text{SG}} = \mathcal{M}(x, s),
\end{equation}
where the output $\mathbf{y}^{\text{SG}}$ is a schema-aligned structured prediction that assigns a value to each field. 

To comprehensively evaluate KIE capabilities, \method{} introduces two complementary evaluation tracks under a schema-guided task formulation.
The \textit{constrained-category KIE track} targets specific application scenarios, where documents are organized by scenario, and each scenario is associated with a small number of predefined extraction schemas. These schemas specify the target fields and output structures, reflecting practical extraction requirements in the real world.

In contrast, the \textit{open-category KIE track} removes scenario-level assumptions by defining extraction schemas at the document level. Rather than relying on shared predefined schemas across documents, each document is paired with a unique schema derived from its own content. This design results in heterogeneous information demands and evaluates whether LMMs can accurately perform KIE over diverse and previously unseen field types.

\begin{table}[t]
\centering
\small

\begin{tabularx}{\linewidth}{>{\raggedright\arraybackslash}p{1.7cm} l c c}
\toprule
\textbf{Domain} & \textbf{Scenario} & \textbf{\#Samples} & \textbf{\#Fields} \\
\midrule

\multirow{4}{*}{\makecell[l]{Business\\Transactions}}
& Commercial        & 620 & 8.56 \\
& Retail            & 347 &4.00  \\
& Catering Services & 212 &5.73  \\
& Accommodation     & 40  & 7.00\\
\midrule

\multirow{4}{*}{\makecell[l]{Public\\Services}}
& Administrative    & 385 &5.83  \\
& Education         & 320 & 3.01 \\
& Postal Label    & 500 &4.00  \\
& Advertisement     & 71  &3.65  \\
\midrule

\multirow{3}{*}{\makecell[l]{Regulated\\Records}}
& Tax-Compliant     & 987 &8.00  \\
& Medical Services  & 240 &4.00  \\
& Nutrition Label   & 750 &8.43  \\
\bottomrule
\end{tabularx}

\caption{Dataset Statistics of the Constrained-Category KIE Track. 
\textbf{\#Field} indicates the averaged field number.}
\label{tab:constrained_kie_statistics}
\end{table}

\subsection{Data Statistics of Different Evaluation Tracks of \method{}}
\label{bench:analysis}
The data statistics of constrained-category and open-category KIE tracks in \method{} are shown in Table~\ref{tab:constrained_kie_statistics} and Table~\ref{tab:open_kie_statistics}, respectively.

The constrained-category track is organized under a hierarchical taxonomy spanning 3 domains and 11 real-world application scenarios, covering business, public service, and regulated document settings. The dataset exhibits substantial variation in both sample scale and schema complexity, with the average number of fields per document ranging from 3 to 9 across scenarios. Business transaction documents show moderate information requirements, while public service documents generally involve more compact schemas. In contrast, regulated records contain more fields on average. This distribution reflects realistic differences across application scenarios and enables fine-grained evaluation of model robustness under varying scenarios.

The open-category track organizes documents by language and document type. While both languages encompass the same set of document types, English documents consistently exhibit more complex schemas. This difference is particularly pronounced for Form and Invoice documents, in which the average number of fields exceeds 15. These statistics indicate systematic variations in information density across languages and document types, further highlighting the heterogeneous extraction requirements inherent in open-category KIE.

\begin{table}[t]
\centering
\small

\begin{tabularx}{\linewidth}{>{\raggedright\arraybackslash}p{1.7cm} X c c}
\toprule
\textbf{Language} & \textbf{Type} & \textbf{\#Samples} & \textbf{\#Fields} \\
\midrule

\multirow{4}{*}{\makecell[l]{Chinese}}
& Receipt   & 202 & 9.88 \\
& Form      & 208 & 17.75 \\
& Invoice   & 207 & 12.11 \\
& Contract  & 208 & 17.38 \\
\midrule

\multirow{4}{*}{\makecell[l]{English}}
& Receipt   & 207 & 13.61 \\
& Form      & 210 & 15.90 \\
& Invoice   & 212 & 17.36 \\
& Contract  & 207 & 17.93 \\
\bottomrule
\end{tabularx}

\caption{Dataset statistics of the Open-Category KIE Track. \textbf{\#Field} indicates the averaged field number.}
\label{tab:open_kie_statistics}
\end{table}

\begin{table*}[t]
\centering
\small
\begin{tabular}{lcccccr}
\toprule
\multirow{2}{*}{\textbf{Benchmark}}
& \multirow{2}{*}{\textbf{LMM-Ready}}
& \multicolumn{2}{c}{\textbf{Track}}
& \multicolumn{2}{c}{\textbf{Taxonomy}}
& \multirow{2}{*}{\textbf{Size}} \\
\cmidrule(lr){3-4}\cmidrule(lr){5-6}
& 
& \textbf{Constrained}
& \textbf{Open}
& \textbf{\#Domains}
& \textbf{\#Scenarios}
& \\
\midrule
OCRBench~\cite{liu2024ocrbench}
& \cmark
& \cmark & \xmark
& -- & 3 & 200 \\
DocILE~\cite{vsimsa2023docile}
& \xmark
& \cmark & \xmark
& 1 & -- & 600 \\
OCRBenchV2~\cite{fu2024ocrbench}
& \cmark
& \cmark & \xmark
& -- & 6 & 800 \\
RealKIE~\cite{townsend2024realkie}
& \xmark
& \cmark & \xmark
& 1 & 5 & 1{,}867 \\
CC-OCR~\cite{yang2025cc}
& \cmark
& \cmark & \cmark
& -- & 6 & 2{,}008 \\
\midrule
\method{}
& \cmark
& \cmark & \cmark
& 3 & 11 & 6{,}133 \\
\bottomrule
\end{tabular}
\caption{Comparison between our \method{} and  Previous KIE Benchmarks. We conduct a detailed comparison of existing KIE benchmarks along four dimensions: support for end-to-end evaluation, task tracks, taxonomy, and dataset size. The symbol ``\textbf{--}'' indicates that the corresponding benchmark does not define or report.}
\label{tab:kie_benchmark_comparison}
\end{table*}

\subsection{Data Curation of \method{}}
\label{bench:data}
In this section, we describe the data curation process of \method{}, which is designed to support a systematic evaluation under the schema-guided KIE formulation.

\textbf{Constrained-Category KIE.}
We curate data from existing benchmarks and organize the constrained-category KIE track according to application scenarios.

Specifically, we collect document images from publicly available document understanding datasets, which are manually assigned to scenarios based on their functional roles and content characteristics. There are three domains (Business Transactions, Public Services, and Regulated Records) that are defined, which comprise 11 application scenarios that span a broad spectrum of real-world document types. Detailed data sources are provided in Appendix~\ref{app:datac}. For each scenario, we identify common information requirements and consolidate key fields into scenario-predefined schemas. When annotations from source datasets are compatible with the schema-guided KIE formulation, they are mapped to the corresponding schema fields; otherwise, the documents are re-annotated accordingly. The detailed curation procedures for this track are described in Appendix~\ref{app:ann}.
To ensure high annotation quality for the constrained-category KIE track, we employ a multi-stage quality control process. For converted annotations, we manually verify field-level semantic consistency with the scenario schema. For re-annotated samples, we conduct model-assisted verification by leveraging advanced LMMs and resolving any discrepancies through human review.

\textbf{Open-Category KIE.} We categorize documents in the open-category KIE track along two dimensions, language and document type, resulting in a collection that spans 2 languages (English and Chinese) and 4 representative document types (receipt, form, invoice, and contract).

We collect documents via a document reconstruction pipeline grounded in real-world document examples.
For each document type, we first curate a small set of representative documents from practical scenarios and include them as in-context examples in the prompt.
The LMM is prompted conditioned on these examples to generate a document description that characterizes the content elements that typically appear in such documents.
Based on this description, we further prompt an LLM to generate executable HTML code that instantiates the described content and layout, explicitly defining the document structure and the spatial arrangement of textual elements.
Rendering the HTML code yields the final document images used in this track. To further enhance visual realism, we subsequently introduce lightweight noise that reflects real-world conditions. 
More curation details are provided in Appendix~\ref{app:open_curation}, while additional authenticity analyses are presented in Appendix~\ref{app:open_analysis}.
To get the ground truth key-value pairs, we follow~\citet{yang2025cc} to annotate documents in the open-category KIE track.
Each document is processed with optical character recognition to obtain textual elements. Then we correct the OCR results, identify key information values, and organize them into a hierarchical extraction schema based on semantic relationships, such as grouping and containment.

\subsection{Benchmark Comparison}
\label{bench:comparsion}
We compare \method{} with representative KIE benchmarks in Table~\ref{tab:kie_benchmark_comparison} along four key dimensions: (i) support for end-to-end LMM evaluation, (ii) coverage of extraction tracks, (iii) document taxonomy design, and (iv) dataset scale.

As shown in Table~\ref{tab:kie_benchmark_comparison}, most existing KIE benchmarks exhibit notable limitations when applied to LMM evaluation. Specifically, DocILE and RealKIE depend on OCR annotations, rendering them unsuitable for end-to-end LMM evaluation. While OCRBench and OCRBenchV2 enable end-to-end evaluation, they mainly target constrained-category KIE via a question-answering formulation and offer either coarse-grained or no explicit document taxonomy. In contrast, \method{} introduces a hierarchical document taxonomy encompassing 3 domains and 11 application scenarios, facilitating fine-grained analyses on different domains and scenarios. Moreover, \method{} unifies constrained-category and open-category KIE tracks under a schema-guided formulation and scales the dataset to 6,133 documents, resulting in a more comprehensive and realistic benchmark for LMM-based KIE evaluation.

\section{Evaluation Protocol}
In this section, we describe the evaluation protocol of \method{}, including the evaluation metrics, baseline models, and implementation details.
\begin{table*}[t]
\centering
\begingroup
\small
\resizebox{\textwidth}{!}{%
\begin{tabular}{l*{11}{r}c}
\toprule
\multicolumn{1}{l}{\multirow{2}{*}[-0.4ex]{\textbf{Method}}}
& \multicolumn{4}{c}{\textbf{Business Transactions}}
& \multicolumn{4}{c}{\textbf{Public Services}}
& \multicolumn{3}{c}{\textbf{Regulated Records}}
& \multicolumn{1}{c}{\multirow{2}{*}[-0.4ex]{\textbf{Avg.}}}
\\
\cmidrule(lr){2-5}\cmidrule(lr){6-9}\cmidrule(lr){10-12}
& \multicolumn{1}{c}{Ret.}
& \multicolumn{1}{c}{Cat.}
& \multicolumn{1}{c}{Com.}
& \multicolumn{1}{c}{Acco.}
& \multicolumn{1}{c}{Post.}
& \multicolumn{1}{c}{Admi.}
& \multicolumn{1}{c}{Edu.}
& \multicolumn{1}{c}{Adv.}
& \multicolumn{1}{c}{Tax.}
& \multicolumn{1}{c}{Med.}
& \multicolumn{1}{c}{Nut.}
& \\
\midrule
\multicolumn{13}{c}{{\cellcolor[rgb]{0.957,0.957,0.957}}\textbf{Close-source LMMs}} \\
\midrule
Claude-Sonnet-4.5    & 69.24 & 73.65 & 45.92 & 69.29 & 79.45 & 69.20 & 39.46 & 85.06 & 67.79 & 63.96 & 67.24 & 66.39 \\
GPT-4o               & 75.79 & 79.33 & 49.67 & 67.86 & 82.95 & 64.26 & 47.77 & 73.56 & 75.33 & 67.57 & 76.51 &69.15  \\
GPT-5                & 71.90& 71.33 & 45.94 & 69.29 & 80.95 & 62.79 & 50.26 & 75.48 & 72.05 & 67.71& 75.83 & 67.59 \\
Qwen-VL-Max          & 86.67 & 84.68 & 49.33 &76.07& 84.85 & 79.96 & 82.46 & 82.76 & 77.94 & 75.62 & 75.14& 77.77\\
Qwen3-VL-Plus        & 89.12 & 86.45 & 50.57 & 77.86 & 83.55 & 78.91 & 87.44 & 83.14 & 78.93 & 76.46 & 89.72 & 80.20 \\
Gemini-3-Pro         & 89.27 & 83.54 & 53.87     & 81.43    & 90.74 & 84.32     & 80.71     & 90.54     & 81.54 & 77.60     & 92.47 & 82.37     \\
\midrule
\multicolumn{13}{c}{{\cellcolor[rgb]{0.957,0.957,0.957}}\textbf{Open-source LMMs}} \\
\midrule
SmolVLM2-2.2B       &17.80&57.22&28.52&18.92 &34.98& 8.63&3.87&27.16&34.97 &18.00&10.51 &23.69\\
Gemma-3-12B         &54.32&68.73&30.68 &53.93 &72.30 &37.14&17.56&36.78&56.81&48.54 &52.20&48.09\\
InternVL3.5-8B       & 77.23 & 72.23 & 41.64 & 67.14 & 73.90 & 72.10 & 63.08 & 73.08 & 72.81 & 65.94& 65.53 & 67.70\\
Ministral-3-8B      &6.47&57.01& 26.24&43.39&75.80& 32.37&29.54&19.53 &65.85 &51.46&6.17&37.62\\
MiniCPM-V4.5-8B      & 68.37 & 74.91 & 45.91 & 67.76 & 83.60 & 71.89 & 63.21 & 79.31 & 69.57 & 70.00& 44.18 & 67.15 \\
GLM-4.1V-9B          & 62.32 & 77.33 & 46.67 & 76.43 & 76.46 & 78.75 & 65.83 & 81.99 & 70.68 & 72.08 & 60.47 & 69.91 \\
Kimi-VL-A3B          & 86.49 & 63.12 & 42.87 & 61.79 & 83.40 & 68.72 & 80.37 & 72.41 & 77.80& 64.93 & 25.72 & 66.14\\
MiMo-VL-7B-RL & 77.45 & 77.66 & 49.01 & 67.50 & 76.70 & 76.81 & 76.65 & 66.28 & 74.86 & 68.70 & 46.10 &68.88 \\
Qwen3-VL-8B          & 88.26 & 86.23 & 51.14 & 71.07 & 83.80 & 78.90 & 87.39 & 81.23 & 78.74 & 77.19 & 86.36 & 79.12 \\
\bottomrule
\end{tabular}%
}
\caption{Overall Performance for Representative LMMs on the Constrained-Category KIE Track of \method{}. Detailed model information is provided in Appendix~\ref{app:more}.}
\label{tab:overall_public}
\endgroup
\end{table*}

\textbf{Baselines.} We evaluate a diverse set of representative LMMs as baselines, covering both proprietary and open-source models with varying model scales, architectures, and training strategies.
Specifically, the baselines include Gemini-3-Pro, Qwen-VL family~\cite{bai2025qwen2}, GPT family, Claude-Sonnet-4.5, SmolVLM2-2.2B~\cite{marafioti2025smolvlm}, Ministral-3-8B~\cite{mistral2025mistral3},
InternVL3.5-8B~\cite{wang2025internvl3}, GLM-4.1V-9B~\cite{glm45vglm41vthinking}, MiMo-VL-7B-RL~\cite{mimovltechnicalreport}, MiniCPM-V-4.5-8B~\cite{yu2025minicpm}, and Kimi-VL-A3B~\cite{team2025kimi}. More details are provided in Appendix~\ref{app:more}. 

\textbf{Evaluation Metrics.}
We evaluate KIE performance using a field-level F1 score, following prior work~\cite{yang2025cc,kim2022ocr}.
Each schema-defined field is evaluated independently. A prediction is considered correct only if the extracted value exactly matches the corresponding ground-truth annotation; any character-level mismatch is treated as an error.
We release the implementation of all normalization rules and metrics used during model evaluation as a part of our proposed \method{}.

\textbf{Implementation Details.}
For all evaluated LMMs, we adopt a unified prompt template across evaluation tracks to ensure fair comparison; the full prompt is provided in Appendix~\ref{app:prompt}. All document images are resized such that the total number of pixels does not exceed 1{,}605{,}632. Proprietary models are accessed via their official SDKs, while all open-source models are deployed using vLLM with Flash-Attention. We set the temperature to 0 for all models to eliminate sampling variability.

\section{Results and Analysis}
This section presents a comprehensive evaluation of representative LMMs on \method{}, with in-depth analyses across diverse application scenarios, document types, and languages.

\begin{table*}[t]
\centering
\small
\setlength{\tabcolsep}{4pt}
\begin{tabular}{lcccccccc c}
\toprule
\multirow{2}{*}{\textbf{Method}} 
& \multicolumn{4}{c}{\textbf{Chinese}} 
& \multicolumn{4}{c}{\textbf{English}} 
& \multirow{2}{*}{\textbf{Avg.}} \\
\cmidrule(lr){2-5}\cmidrule(lr){6-9}
& Receipt & Form & Invoice & Contract 
& Receipt & Form & Invoice & Contract & \\
\midrule

\multicolumn{10}{c}{\cellcolor[rgb]{0.95,0.95,0.95}\textbf{Closed-source LMMs}} \\
\midrule
Claude-Sonnet-4.5   & 37.59 & 31.20 & 29.12 & 20.88 & 41.64 & 49.22 & 44.18 & 63.87 & 39.71 \\
GPT-4o              & 39.66 & 30.82 & 35.31 & 16.67 & 78.51 & 48.09 & 57.08 & 67.35 & 46.69 \\
GPT-5               & 39.61 & 31.97 & 37.64 & 17.47 & 80.64 & 50.09 & 57.50 & 64.24 & 47.39 \\
Qwen-VL-Max         & 69.94 & 69.75 & 66.62 & 69.68 & 70.76 & 65.52 & 70.80 & 76.57 & 69.95 \\
Qwen3-VL-Plus       & 71.14 & 69.87 & 69.59 & 70.95 & 68.27 & 66.82 & 73.45 & 74.81 & 70.61 \\
Gemini-3-Pro        & 78.50 & 79.17 & 78.63 & 75.96 & 93.57 & 80.41 & 87.06 & 79.90 & 81.65 \\

\midrule
\multicolumn{10}{c}{\cellcolor[rgb]{0.95,0.95,0.95}\textbf{Open-source LMMs}} \\
\midrule
SmolVLM2-2.2B &5.74 &1.51   &3.05 &2.15   & 36.49 &11.64  &19.73  &22.36  &12.83  \\ 
Gemma-3-12B         & 25.71 & 14.17 & 17.36 &  9.42 & 76.24 & 38.32 & 30.23 & 35.77 & 30.90 \\
InternVL3.5-8B      & 54.26 & 37.83 & 40.97 & 39.43 & 79.46 & 43.64 & 49.41 & 50.06 & 49.38 \\
Ministral-3-8B &25.21 &35.80   &24.42 &26.18   &58.54 &49.01  &55.66  &57.82  &41.58  \\
MiniCPM-V4.5-8B     & 52.33 & 42.69 & 44.54 & 39.46 & 72.41 & 47.46 & 56.44 & 65.49 & 52.60 \\
GLM-4.1V-9B         & 55.19 & 56.63 & 55.52 & 54.68 & 63.83 & 54.66 & 59.31 & 68.70 & 58.56 \\
Kimi-VL-A3B         & 62.98 & 52.85 & 50.72 & 54.94 & 73.91 & 55.96 & 52.67 & 64.64 & 58.58 \\
MiMo-VL-7B-RL       & 57.63 & 56.60 & 51.93 & 65.90 & 71.34 & 52.83 & 57.64 & 70.55 & 60.55 \\
Qwen3-VL-8B         & 68.93 & 64.70 & 63.38 & 65.59 & 71.40 & 61.46 & 67.34 & 76.08 & 67.36 \\
\bottomrule
\end{tabular}
\caption{Overall Performance of Representative LMMs on the Open-Category KIE Track of \method{}. Detailed model information is provided in Appendix~\ref{app:more}.}
\label{tab:overall_open}
\end{table*}

\subsection{Overall Performance on Different KIE Tracks in \method{}}
This subsection presents the performance of various models on both the Constrained Category and Open Category KIE tracks in \method{}.

\textbf{KIE for Constrained Category.} Table~\ref{tab:overall_public} displays the performance of representative LMMs on the constrained-category KIE track of \method{}, with results across different scenarios.

A clear performance gap is observed between closed-source and open-source LMMs, underscoring the significance of strong multimodal alignment in KIE. Among the proprietary LMMs, Gemini-3-Pro achieves the best overall performance, followed by Qwen3-VL-Plus and Qwen-VL-Max, all of which exhibit stable and consistently strong results across business, public service, and regulated document scenarios. In contrast, most open-source models underperform, particularly in the public service and regulated document categories, where long text spans, domain-specific terminology, and implicit field dependencies impose higher demands on visual understanding. Despite similar model scales, open-source LMMs show substantial performance variation, suggesting that scale alone does not account for the observed differences. Notably, Qwen3-VL-8B emerges as a strong open-source baseline, approaching the performance of some proprietary LMMs on average, but still shows degradation in the most challenging categories.

\textbf{KIE for Open Category.}
Table~\ref{tab:overall_open} presents the overall performance of representative LMMs on the Open-Category KIE track.

In general, closed-source models outperform their open-source counterparts, with Gemini-3-Pro achieving the highest average F1 score and demonstrating robust performance across both languages and document types. We observe a consistent and substantial performance drop from English to Chinese across most models, highlighting a persistent cross-lingual robustness challenge in open-category KIE. This disparity is likely influenced by the dominance of English document distributions in vision-language pretraining. Additionally, the compact glyph structure, high character density, and absence of explicit word boundaries in Chinese texts further complicate visual recognition and boundary alignment, which negatively impacts KIE performance.
The performance varies significantly across different document types. Specifically, receipt documents consistently present the least difficulty for all models, yielding the highest scores due to their relatively regular layouts and well-defined key–value structures. In contrast, form documents show the greatest challenge, especially for smaller and open-source models. This suggests that the fragmented layouts and flexible field boundaries in forms make field localization and extraction more prone to errors.
Invoice and contract documents show intermediate performance but exhibit different behaviors across languages: while contracts perform reasonably well in English, Chinese contracts remain particularly challenging, especially for smaller-scale LMMs.

\subsection{Faithfulness Analysis of LMMs in KIE}
We further investigate the faithfulness of LMMs in KIE by examining whether the predicted field values are grounded in the input document image, rather than arising from visual perception errors or hallucinated content.
Specifically, we conduct this analysis on the open-category KIE track, which benefits from validated OCR results and thus enables a reliable estimation of hallucination behaviors in LMMs. An extracted value is considered faithful if it can be located in the OCR results of the corresponding document.

\begin{figure}[t]
    \centering
    \begin{subfigure}[t]{0.48\linewidth}
        \centering
        \includegraphics[width=\linewidth]{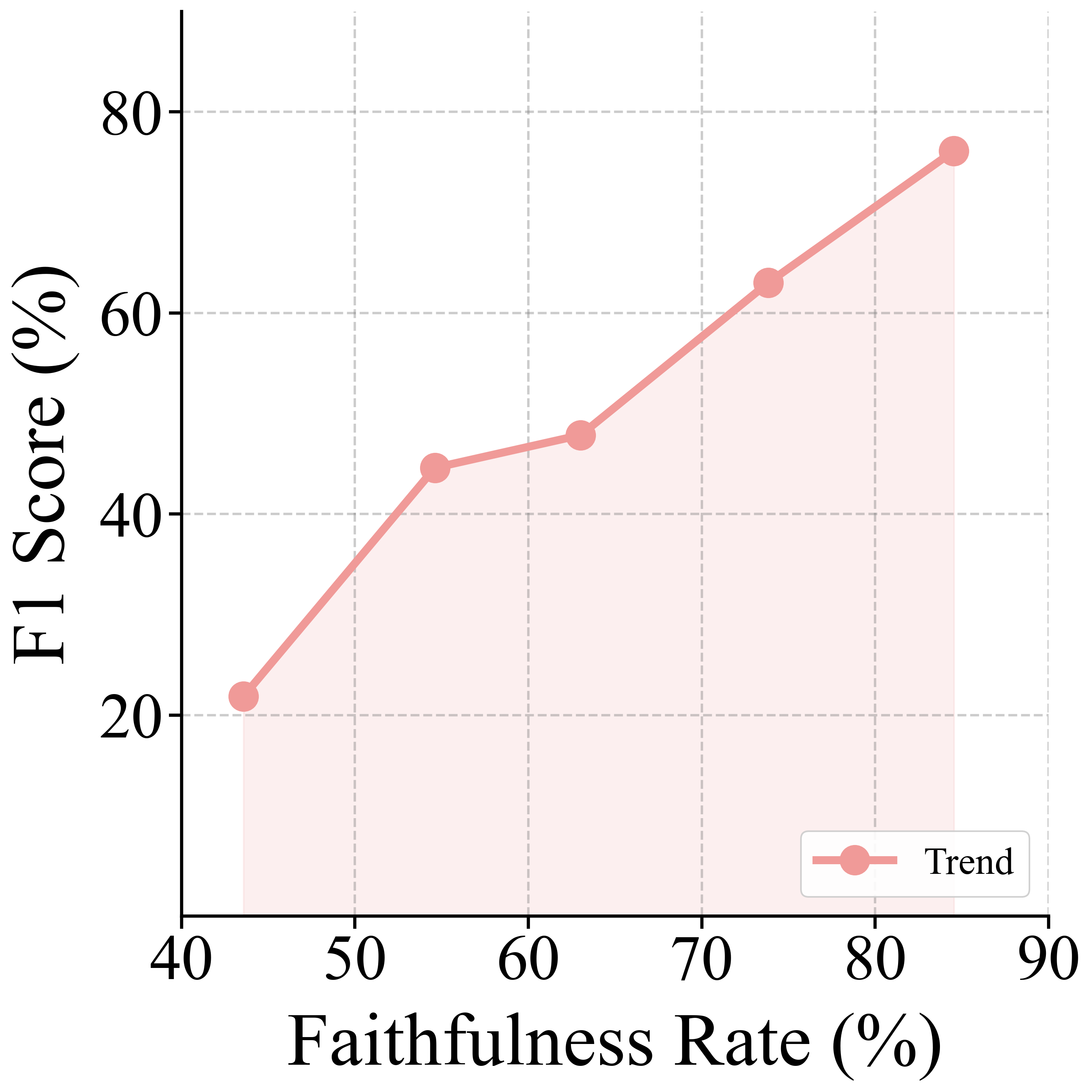}
        \caption{Correlation between the Faithfulness Rate and F1 Score. }
        \label{fig:ff_correlation}
    \end{subfigure}
    \hfill
    \begin{subfigure}[t]{0.48\linewidth}
        \centering
        \includegraphics[width=\linewidth]{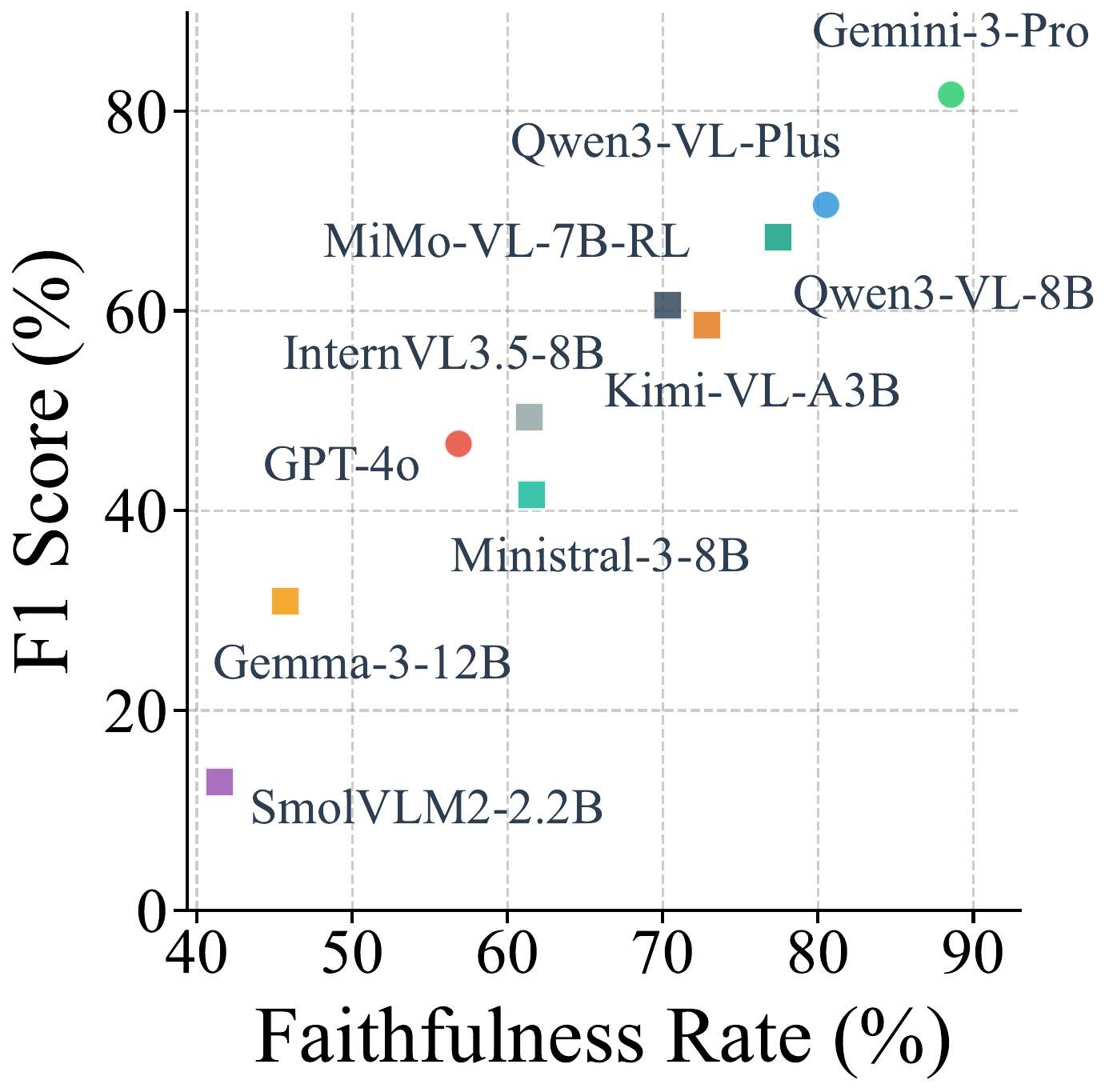} %
        \caption{F1 Score and Faithfulness Rate across Different LMMs. }
        \label{fig:ff_lmm}
    \end{subfigure}
    \caption{Faithfulness Analysis of LMMs in KIE\label{fig:ff}. We analyzed the relationship between the faithfulness and the extraction performance of LMMs in KIE.}
\end{figure}

As shown in Figure~\ref{fig:ff_correlation}, when averaged across all evaluated models, higher F1 scores are generally associated with higher faithfulness rates. This positive correlation indicates that stronger KIE performance is typically accompanied by more reliable grounding of predicted field values in the document content. Figure~\ref{fig:ff_lmm} further provides a fine-grained analysis across different LMMs and reveals notable variations among models. Importantly, even among LMMs exhibiting comparable faithfulness rates, we observe substantial disparities in F1 scores. This observation suggests that while faithfulness constitutes a necessary prerequisite for conducting accurate KIE results, it is not sufficient on its own to guarantee strong extraction performance. The primary reason is that, even when models exhibit similar grounding behaviors, they can still differ substantially in their ability to capture field semantics, precisely delineate field boundaries, and correctly select the target instance among multiple candidates.

\subsection{Typical Error Analysis}
\label{error_analysis}
\begin{figure}
    \centering
    \includegraphics[width=\linewidth]{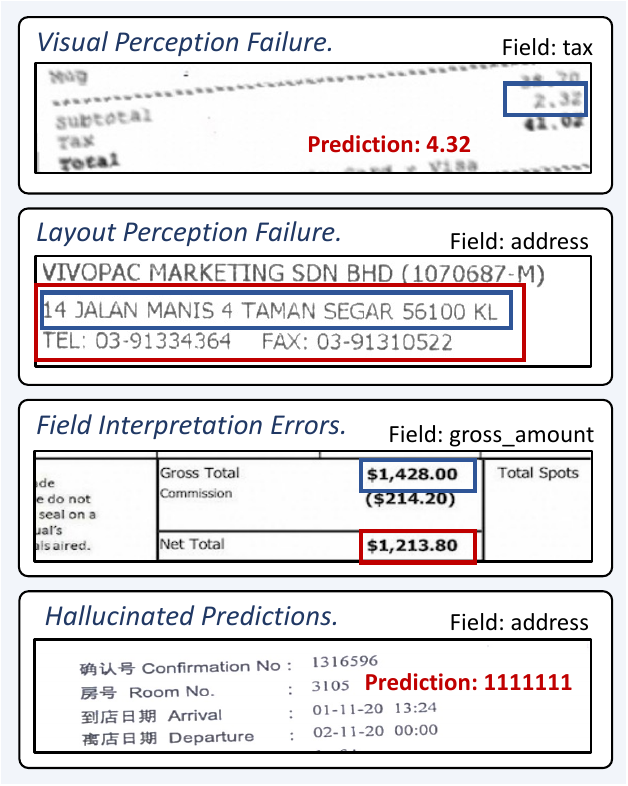}
\caption{Typical Error Cases of LMMs in KIE. Blue boxes indicate the ground truth, while red boxes denote the model predictions.}
\label{fig:error_analysis}
\end{figure}

To better understand the remaining performance gaps, we conduct a typical error analysis. As illustrated in Figure~\ref{fig:error_analysis}, we identify several recurring error patterns, which shed light on the systematic limitations of current LMMs in key information extraction. A more detailed behavioral analysis of LMMs in KIE is provided in Appendix~\ref{app:behavior}.

\textbf{Visual Perception Failure.}
In this error category, the model fails at the level of visual text recognition. For the tax field, the ground-truth value shown in the document is 2.32, while the model predicts 4.32. Such errors mainly arise from imperfect visual perception (e.g., digit confusion or misrecognition), rather than from semantic misunderstanding of the field.

\textbf{Layout Perception Failure.}
These errors occur when the model correctly recognizes textual content but fails to associate fields with their corresponding spatial regions. For example, when extracting an address field, the model selects an incorrect nearby line. This error pattern indicates deficiencies in layout understanding, where the model struggles to align a target field with the correct text span when layout boundaries are ambiguous.

\textbf{Field Interpretation Errors.}
This error type refers to cases where the model successfully localizes relevant values but assigns them to incorrect fields. In our examples, the model confuses the semantic roles of fields, mapping the net total to the gross amount. Such errors reveal limitations in fine-grained field understanding, especially in scenarios requiring domain-specific knowledge or nuanced distinctions between closely related fields.

\textbf{Hallucinated Predictions.}
Hallucinated errors correspond to predictions that are not grounded in the document content. In the illustrated case, the model outputs an address value that does not appear anywhere in the input image.
Unlike perception or interpretation errors, these predictions are fabricated without visual or textual evidence. This error pattern highlights the challenge of maintaining faithfulness in KIE, particularly when target fields are missing and incomplete.
\section{Conclusion}
We present \method{}, a unified benchmark for evaluating KIE in realistic and heterogeneous document settings. Experiments on representative LMMs show that existing models still face challenges when confronted with diverse extraction requirements, long-tail fields, and complex document layouts. We hope that \method{} can serve as a reliable and practical evaluation to facilitate the development of more robust and faithful document understanding models.


\section*{Limitations}
While \method{} offers a systematic and comprehensive evaluation of the KIE capabilities of current LMMs, it is restricted to single-page or short-document scenarios and does not address long-document KIE.
This limitation arises because long-document KIE typically depends on auxiliary components (\textit{e.g.}, page retrieval and cross-page aggregation), whose performance often dominates overall results under the current context-length constraints of LMMs. As a consequence, isolating and evaluating the intrinsic KIE ability of LMMs in long-document settings becomes challenging.
To facilitate a more controlled, interpretable, and focused assessment of the core KIE capabilities of LMMs, \method{} deliberately confines its evaluation scope to short documents.

\section*{Ethical Considerations}
All documents included in \method{} are sourced either from publicly available datasets or generated through carefully designed privacy-preserving pipelines. Throughout the construction of the benchmark, strict measures are taken to ensure that no personal, sensitive, or identifiable information is collected, processed, or disclosed.
A portion of the dataset is annotated by the collaborators of this work. No external annotators, crowdworkers, or paid participants are involved; all annotations are conducted voluntarily by contributors with relevant expertise. Before the annotation process, all annotators are fully informed of the research objectives, the intended use of the annotated data, and the evaluation protocols of the benchmark. To ensure consistency and reliability, we provide comprehensive annotation guidelines that clearly specify task definitions, annotation formats, and operational constraints.
Importantly, the annotation process for key information extraction from documents does not constitute human-subjects research. It does not involve interaction with, or data collection from, external individuals, and therefore does not raise concerns related to human participant protection or personal data usage.

\section*{Acknowledgments}
This work is partly supported by the National Natural Science Foundation of China (No. 62576082 and No. 62461146205) and Alibaba Innovative Research Program.


\bibliography{custom}

@article{rombach2025deep,
  title={Deep learning based key information extraction from business documents: Systematic literature review},
  author={Rombach, Alexander Michael and Fettke, Peter},
  journal={ACM Computing Surveys},
  pages={1--37},
  year={2025},
  url ={https://dl.acm.org/doi/full/10.1145/3749369}
}

@online{mistral2025mistral3,
  title        = {Introducing Mistral 3},
  author       = {{Mistral AI}},
  year         = {2025},
  month        = dec,
  day          = {2},
  url          = {https://mistral.ai/news/mistral-3},
  urldate      = {2026-01-03},
  organization = {Mistral AI}
}

@article{aiello2002document,
  title={Document Understanding for a Broad Class of Documents},
  author={Aiello, Marco and Monz, C and Todoran, L and Worring, M},
  journal={International Journal on Document Analysis and Recognition},
  pages={1--16},
  year={2002},
  url = {https://staff.fnwi.uva.nl/c.monz/html/publications/ijdar.pdf}
}

@article{marafioti2025smolvlm,
  title={Smolvlm: Redefining small and efficient multimodal models},
  author={Marafioti, Andr{\'e}s and Zohar, Orr and Farr{\'e}, Miquel and Noyan, Merve and Bakouch, Elie and Cuenca, Pedro and Zakka, Cyril and Allal, Loubna Ben and Lozhkov, Anton and Tazi, Nouamane and others},
  journal={ArXiv preprint},
  year={2025},
  url = {https://arxiv.org/abs/2504.05299},
  volume = {abs/2504.05299},
}

@article{ge2024scaling,
  title={Scaling synthetic data creation with 1,000,000,000 personas},
  author={Ge, Tao and Chan, Xin and Wang, Xiaoyang and Yu, Dian and Mi, Haitao and Yu, Dong},
  journal={ArXiv preprint},
  year={2024},
  url = {https://arxiv.org/abs/2406.20094},
  volume = {abs/2406.20094},
}

@article{townsend2024realkie,
  title={RealKIE: Five Novel Datasets for Enterprise Key Information Extraction},
  author={Townsend, Benjamin and May, Madison and Mackowiak, Katherine and Wells, Christopher},
  journal={ArXiv preprint},
  year={2024},
  url = {https://arxiv.org/abs/2403.20101},
  volume = {abs/2403.20101},
}

@article{bai2025qwen3vltechnicalreport,
      title={Qwen3-VL Technical Report}, 
      author={Shuai Bai and Yuxuan Cai and Ruizhe Chen and Keqin Chen and Xionghui Chen and others},
      year={2025},
      journal={ArXiv preprint},
      volume = {abs/2511.21631},
      url = {https://arxiv.org/abs/2511.21631},
      
}

@article{wang2025internvl3,
  title={Internvl3. 5: Advancing open-source multimodal models in versatility, reasoning, and efficiency},
  author={Wang, Weiyun and Gao, Zhangwei and Gu, Lixin and Pu, Hengjun and Cui, Long and Wei, Xingguang and Liu, Zhaoyang and Jing, Linglin and Ye, Shenglong and Shao, Jie and others},
  journal={ArXiv preprint},
  year={2025},
  url = {https://arxiv.org/abs/2508.18265},
  volume = {abs/2508.18265}
}

@article{glm45vglm41vthinking,
      title={GLM-4.5V and GLM-4.1V-Thinking: Towards Versatile Multimodal Reasoning with Scalable Reinforcement Learning}, 
      author={V Team and Wenyi Hong and Wenmeng Yu and Xiaotao Gu and Guo Wang and Guobing Gan and Haomiao Tang and others},
      year={2025},
      journal={ArXiv preprint},
      url = {https://arxiv.org/abs/2507.01006},
      volume = {abs/2507.01006},
}

@article{mimovltechnicalreport,
      title={MiMo-VL Technical Report}, 
      author={Core Team and Zihao Yue and Zhenru Lin and Yifan Song and Weikun Wang and Shuhuai Ren and Shuhao Gu and Shicheng Li and Peidian Li and others},
      year={2025},
      journal={ArXiv preprint}, 
      url = {https://arxiv.org/abs/2506.03569},
      volume = {abs/2506.03569},
}

@article{team2025kimi,
  title={Kimi-vl technical report},
  author={Team, Kimi and Du, Angang and Yin, Bohong and Xing, Bowei and Qu, Bowen and Wang, Bowen and Chen, Cheng and Zhang, Chenlin and Du, Chenzhuang and Wei, Chu and others},
  journal={ArXiv preprint},
  year={2025},
  url = {https://arxiv.org/abs/2504.07491},
  volume = {abs/2504.07491},
}

@inproceedings{kuang2023visual,
  title={Visual information extraction in the wild: practical dataset and end-to-end solution},
  author={Kuang, Jianfeng and Hua, Wei and Liang, Dingkang and Yang, Mingkun and Jiang, Deqiang and Ren, Bo and Bai, Xiang},
  booktitle={Proceedings of ICDAR},
  pages={36--53},
  year={2023},
  url = {https://link.springer.com/chapter/10.1007/978-3-031-41731-3_3}
}

@inproceedings{vsimsa2023docile,
  title={Docile benchmark for document information localization and extraction},
  author={{\v{S}}imsa, {\v{S}}t{\v{e}}p{\'a}n and {\v{S}}ulc, Milan and U{\v{r}}i{\v{c}}{\'a}{\v{r}}, Michal and Patel, Yash and Hamdi, Ahmed and Koci{\'a}n, Mat{\v{e}}j and Skalick{\`y}, Maty{\'a}{\v{s}} and Matas, Ji{\v{r}}{\'\i} and Doucet, Antoine and Coustaty, Micka{\"e}l and others},
  booktitle={International Conference on Document Analysis and Recognition},
  pages={147--166},
  year={2023},
  url = {https://link.springer.com/chapter/10.1007/978-3-031-41679-8_9},
  organization={Springer}
}

@inproceedings{laatiri2023information,
  title={Information redundancy and biases in public document information extraction benchmarks},
  author={Laatiri, Seif and Ratnamogan, Pirashanth and Tang, Jo{\"e}l and Lam, Laurent and Vanhuffel, William and Caspani, Fabien},
  booktitle={Proceedings of ICDAR},
  pages={280--294},
  year={2023},
  url = {https://link.springer.com/chapter/10.1007/978-3-031-41682-8_18}
}

@inproceedings{bai2025longbench,
  title={Longbench v2: Towards deeper understanding and reasoning on realistic long-context multitasks},
  author={Bai, Yushi and Tu, Shangqing and Zhang, Jiajie and Peng, Hao and Wang, Xiaozhi and Lv, Xin and Cao, Shulin and Xu, Jiazheng and Hou, Lei and Dong, Yuxiao and others},
  booktitle={Proceedings of ACL},
  pages={3639--3664},
  year={2025},
  url = {https://aclanthology.org/2025.acl-long.183/}
  
}

@article{comanici2025gemini,
  title={Gemini 2.5: Pushing the frontier with advanced reasoning, multimodality, long context, and next generation agentic capabilities},
  author={Comanici, Gheorghe and Bieber, Eric and Schaekermann, Mike and Pasupat, Ice and Sachdeva, Noveen and Dhillon, Inderjit and Blistein, Marcel and Ram, Ori and Zhang, Dan and Rosen, Evan and others},
  journal={ArXiv preprint},
  url = {https://arxiv.org/abs/2507.06261},
  volume = {abs/2507.06261},
  year={2025}
}

@article{gbada2025deep,
  title={Deep learning approaches for information extraction from visually rich documents: Datasets, challenges and methods},
  author={Gbada, Hamza and Kalti, Karim and Mahjoub, Mohamed Ali},
  journal={International Journal on Document Analysis and Recognition (IJDAR)},
  pages={121--142},
  year={2025},
  url = {https://link.springer.com/article/10.1007/s10032-024-00493-8#citeas}
}

@inproceedings{vafaie2025end,
  title={End-to-end Information Extraction from Archival Records with Multimodal Large Language Models},
  author={Vafaie, Mahsa and Hertling, Sven and Banse-Strobel, Inger and Dubout, Kevin and Sack, Harald},
  booktitle={Proceedings of CIKM},
  pages={6075--6083},
  year={2025},
  url = {https://dl.acm.org/doi/abs/10.1145/3746252.3761503}
}

@article{bai2025qwen2,
  title={Qwen2.5-vl technical report},
  author={Bai, Shuai and Chen, Keqin and Liu, Xuejing and Wang, Jialin and Ge, Wenbin and Song, Sibo and Dang, Kai and Wang, Peng and Wang, Shijie and Tang, Jun and others},
  journal={ArXiv preprint},
  year={2025},
  url = {https://arxiv.org/abs/2502.13923},
  volume = {abs/2502.13923},
}

@article{tang1994document,
  title={Document processing for automatic knowledge acquisition},
  author={Tang, Yuan Yan and De Yan, Chang and Suen, Ching Y.},
  journal={IEEE transactions on Knowledge and Data Engineering},
  pages={3--21},
  year={1994},
  url = {https://ieeexplore.ieee.org/abstract/document/273022}
}

@article{mao2003document,
  title={Document structure analysis algorithms: a literature survey},
  author={Mao, Song and Rosenfeld, Azriel and Kanungo, Tapas},
  journal={Document recognition and retrieval X},
  pages={197--207},
  year={2003},
  url = {https://www.spiedigitallibrary.org/conference-proceedings-of-spie/5010/0000/Document-structure-analysis-algorithms-a-literature-survey/10.1117/12.476326.short?tab=ArticleLinkCited}
}

@article{cesarini2002informys,
  title={INFORMys: A flexible invoice-like form-reader system},
  author={Cesarini, Francesca and Gori, Marco and Marinai, Simone and Soda, Giovanni},
  journal={IEEE Transactions on Pattern Analysis and Machine Intelligence},
  pages={730--745},
  year={2002},
  url = {https://ieeexplore.ieee.org/abstract/document/689303}
}

@article{ding2024deep,
  title={Deep learning based visually rich document content understanding: A survey},
  author={Ding, Yihao and Han, Soyeon Caren and Lee, Jean and Hovy, Eduard},
  journal={ArXiv preprint},
  year={2024},
  url = {https://arxiv.org/abs/2408.01287},
  volume = {abs/2408.01287}
}

@inproceedings{zhang2023reading,
    title = "Reading Order Matters: Information Extraction from Visually-rich Documents by Token Path Prediction",
    author = "Zhang, Chong  and
      Guo, Ya  and
      Tu, Yi  and
      Chen, Huan  and
      Tang, Jinyang  and
      Zhu, Huijia  and
      Zhang, Qi  and
      Gui, Tao",
    booktitle = "Proceedings of EMNLP",
    year = "2023",
    url = "https://aclanthology.org/2023.emnlp-main.846/",
    pages = "13716--13730",
}

@article{yu2025minicpm,
  title={Minicpm-v 4.5: Cooking efficient mllms via architecture, data, and training recipe},
  author={Yu, Tianyu and Wang, Zefan and Wang, Chongyi and Huang, Fuwei and Ma, Wenshuo and He, Zhihui and Cai, Tianchi and Chen, Weize and Huang, Yuxiang and Zhao, Yuanqian and others},
  journal={ArXiv preprint},
  year={2025},
  volume = {abs/2509.18154},
  url = {https://arxiv.org/abs/2509.18154}
}

@article{lv2023kosmos,
  title={Kosmos-2.5: A multimodal literate model},
  author={Lv, Tengchao and Huang, Yupan and Chen, Jingye and Zhao, Yuzhong and Jia, Yilin and Cui, Lei and Ma, Shuming and Chang, Yaoyao and Huang, Shaohan and Wang, Wenhui and others},
  journal={ArXiv preprint},
  year={2023},
  url = {https://arxiv.org/abs/2309.11419},
  volume = {abs/2309.11419},
}

@inproceedings{nacson2025docvlm,
  title={Docvlm: Make your vlm an efficient reader},
  author={Nacson, Mor Shpigel and Aberdam, Aviad and Ganz, Roy and Ben Avraham, Elad and Golts, Alona and Kittenplon, Yair and Mazor, Shai and Litman, Ron},
  booktitle={Proceedings of CVPR},
  pages={29005--29015},
  year={2025},
  url = {https://openaccess.thecvf.com/content/CVPR2025/html/Nacson_DocVLM_Make_Your_VLM_an_Efficient_Reader_CVPR_2025_paper.html}
}

@article{nasar2018information,
  title={Information extraction from scientific articles: a survey},
  author={Nasar, Zara and Jaffry, Syed Waqar and Malik, Muhammad Kamran},
  journal={Scientometrics},
  pages={1931--1990},
  year={2018},
  url = {https://link.springer.com/article/10.1007/s11192-018-2921-5}
}

@article{oral2020information,
  title={Information extraction from text intensive and visually rich banking documents},
  author={Oral, Berke and Emekligil, Erdem and Arslan, Se{\c{c}}il and Eryiǧit, G{\"u}l{\c{s}}en},
  journal={Information Processing \& Management},
  pages={102361},
  year={2020},
  url = {https://www.sciencedirect.com/science/article/pii/S0306457320308566}
}

@inproceedings{park2019cord,
  title={Cord: a consolidated receipt dataset for post-ocr parsing},
  author={Park, Seunghyun and Shin, Seung and Lee, Bado and Lee, Junyeop and Surh, Jaeheung and Seo, Minjoon and Lee, Hwalsuk},
  booktitle={Workshop on Document Intelligence at NeurIPS 2019},
  year={2019},
  url = {https://openreview.net/forum?id=SJl3z659UH}
}

@article{liu2024ocrbench,
  title={Ocrbench: on the hidden mystery of ocr in large multimodal models},
  author={Liu, Yuliang and Li, Zhang and Huang, Mingxin and Yang, Biao and Yu, Wenwen and Li, Chunyuan and Yin, Xu-Cheng and Liu, Cheng-Lin and Jin, Lianwen and Bai, Xiang},
  journal={Science China Information Sciences},
  pages={220102},
  year={2024},
  url = {https://link.springer.com/article/10.1007/s11432-024-4235-6}
}

@inproceedings{he2023good,
  title={Do-GOOD: towards distribution shift evaluation for pre-trained visual document understanding models},
  author={He, Jiabang and Hu, Yi and Wang, Lei and Xu, Xing and Liu, Ning and Liu, Hui and Shen, Heng Tao},
  booktitle={Proceedings of SIGIR},
  pages={569--579},
  year={2023},
  url = {https://dl.acm.org/doi/abs/10.1145/3539618.3591670}
}

@inproceedings{wei2025p2net,
  title={P$^2$Net: Parallel Pointer-based Network for Key Information Extraction with Complex Layouts},
  author={Wei, Kaiwen and Yao, Jie and Zhong, Jiang and Kang, Yangyang and Zhang, Jingyuan and Sun, Changlong and Zhang, Xin and Lv, Fengmao and Jin, Li},
  booktitle={Proceedings of ACL Findings},
  pages={10611--10626},
  year={2025},
  url = {https://aclanthology.org/2025.findings-acl.552/}
}

@article{ebrahimi2022test,
  title={Test-time adaptation for visual document understanding},
  author={Ebrahimi, Sayna and Arik, Sercan O and Pfister, Tomas},
  journal={ArXiv preprint},
  year={2022},
  volume = {abs/2206.07240},
  url = {https://arxiv.org/abs/2206.07240}
}

@inproceedings{huang2022layoutlmv3,
  title={Layoutlmv3: Pre-training for document ai with unified text and image masking},
  author={Huang, Yupan and Lv, Tengchao and Cui, Lei and Lu, Yutong and Wei, Furu},
  booktitle={Proceedings of MM},
  pages={4083--4091},
  year={2022},
  url = {https://dl.acm.org/doi/abs/10.1145/3503161.3548112}
}

@inproceedings{zhang2025dockylin,
  title={Dockylin: A large multimodal model for visual document understanding with efficient visual slimming},
  author={Zhang, Jiaxin and Yang, Wentao and Lai, Songxuan and Xie, Zecheng and Jin, Lianwen},
  booktitle={Proceedings of AAAI},
  pages={9923--9932},
  year={2025},
  url={https://ojs.aaai.org/index.php/AAAI/article/view/33076}
}

@inproceedings{liu2024hrvda,
  title={Hrvda: High-resolution visual document assistant},
  author={Liu, Chaohu and Yin, Kun and Cao, Haoyu and Jiang, Xinghua and Li, Xin and Liu, Yinsong and Jiang, Deqiang and Sun, Xing and Xu, Linli},
  booktitle={Proceedings of CVPR},
  pages={15534--15545},
  year={2024},
  url ={https://openaccess.thecvf.com/content/CVPR2024/html/Liu_HRVDA_High-Resolution_Visual_Document_Assistant_CVPR_2024_paper.html}
}

@article{yu2024texthawk,
  title={Texthawk: Exploring efficient fine-grained perception of multimodal large language models},
  author={Yu, Ya-Qi and Liao, Minghui and Wu, Jihao and Liao, Yongxin and Zheng, Xiaoyu and Zeng, Wei},
  journal={ArXiv preprint},
  year={2024},
  url = {https://arxiv.org/abs/2404.09204},
  volume = {abs/2404.09204},
}

@inproceedings{ye2023ureader,
  title={Ureader: Universal ocr-free visually-situated language understanding with multimodal large language model},
  author={Ye, Jiabo and Hu, Anwen and Xu, Haiyang and Ye, Qinghao and Yan, Ming and Xu, Guohai and Li, Chenliang and Tian, Junfeng and Qian, Qi and Zhang, Ji and others},
  booktitle={Proceedings of EMNLP Findings},
  pages={2841--2858},
  url = {https://aclanthology.org/2023.findings-emnlp.187/},
  year={2023}
}

@article{liu2024textmonkey,
  title={Textmonkey: An ocr-free large multimodal model for understanding document},
  author={Liu, Yuliang and Yang, Biao and Liu, Qiang and Li, Zhang and Ma, Zhiyin and Zhang, Shuo and Bai, Xiang},
  journal={ArXiv preprint},
  year={2024},
  url = {https://arxiv.org/abs/2403.04473},
  volume = {abs/2403.04473},
}

@inproceedings{hu2025mplug,
  title={mplug-docowl2: High-resolution compressing for ocr-free multi-page document understanding},
  author={Hu, Anwen and Xu, Haiyang and Zhang, Liang and Ye, Jiabo and Yan, Ming and Zhang, Ji and Jin, Qin and Huang, Fei and Zhou, Jingren},
  booktitle={Proceedings of ACL},
  pages={5817--5834},
  year={2025},
  url = {https://aclanthology.org/2025.acl-long.291/}
}

@inproceedings{park2024hierarchical,
  title={Hierarchical visual feature aggregation for ocr-free document understanding},
  author={Park, Jaeyoo and Choi, Jin Y and Park, Jeonghyung and Han, Bohyung},
  booktitle={Proceedings of NeurIPS},
  pages={105972--105996},
  year={2024},
  url ={https://proceedings.neurips.cc/paper_files/paper/2024/hash/bf85879363044ca21f7868a3d1b4021c-Abstract-Conference.html}
}

@article{ke2025large,
  title={Large Language Models in Document Intelligence: A Comprehensive Survey, Recent Advances, Challenges, and Future Trends},
  author={Ke, Wenjun and Zheng, Yifan and Li, Yining and Xu, Hengyuan and Nie, Dong and Wang, Peng and He, Yao},
  journal={ACM Transactions on Information Systems},
  pages={1--64},
  year={2025},
  url = {https://dl.acm.org/doi/full/10.1145/3768156}
}

@article{wang2025document,
  title={Document Intelligence in the Era of Large Language Models: A Survey},
  author={Wang, Weishi and Hu, Hengchang and Zhang, Zhijie and Li, Zhaochen and Shao, Hongxin and Dahlmeier, Daniel},
  journal={ArXiv preprint},
  year={2025},
  url = {https://arxiv.org/abs/2510.13366},
  volume = {abs/2510.13366}
}

@article{shi2023exploring,
  title={Exploring ocr capabilities of gpt-4v (ision): A quantitative and in-depth evaluation},
  author={Shi, Yongxin and Peng, Dezhi and Liao, Wenhui and Lin, Zening and Chen, Xinhong and Liu, Chongyu and Zhang, Yuyi and Jin, Lianwen},
  journal={ArXiv preprint},
  year={2023},
  url = {https://arxiv.org/abs/2310.16809},
  volume = {abs/2310.16809},
}

@article{barboule2025survey,
  title={Survey on Question Answering over Visually Rich Documents: Methods, Challenges, and Trends},
  author={Barboule, Camille and Piwowarski, Benjamin and Chabot, Yoan},
  journal={ArXiv preprint},
  year={2025},
  url = {https://arxiv.org/abs/2501.02235},
  volume = {abs/2501.02235},
}

@inproceedings{zhang2025ocr,
  title={Ocr hinders rag: Evaluating the cascading impact of ocr on retrieval-augmented generation},
  author={Zhang, Junyuan and Zhang, Qintong and Wang, Bin and Ouyang, Linke and Wen, Zichen and Li, Ying and Chow, Ka-Ho and He, Conghui and Zhang, Wentao},
  booktitle={Proceedings of ICCV},
  pages={17443--17453},
  url = {https://openaccess.thecvf.com/content/ICCV2025/html/Zhang_OCR_Hinders_RAG_Evaluating_the_Cascading_Impact_of_OCR_on_ICCV_2025_paper.html},
  year={2025}
}

@inproceedings{xu2020layoutlm,
  title={Layoutlm: Pre-training of text and layout for document image understanding},
  author={Xu, Yiheng and Li, Minghao and Cui, Lei and Huang, Shaohan and Wei, Furu and Zhou, Ming},
  booktitle={Proceedings of SIGKDD},
  pages={1192--1200},
  year={2020},
  url = {https://dl.acm.org/doi/10.1145/3394486.3403172}
}

@article{abdallah2024survey,
  title={A survey of recent approaches to form understanding in scanned documents},
  author={Abdallah, Abdelrahman and Eberharter, Daniel and Pfister, Zoe and Jatowt, Adam},
  journal={Artificial Intelligence Review},
  pages={342},
  year={2024},
  publisher={Springer},
  url = {https://link.springer.com/article/10.1007/s10462-024-11000-0}
}

@inproceedings{huang2019icdar2019,
  title={Icdar2019 competition on scanned receipt ocr and information extraction},
  author={Huang, Zheng and Chen, Kai and He, Jianhua and Bai, Xiang and Karatzas, Dimosthenis and Lu, Shijian and Jawahar, CV},
  booktitle={Proceedings of ICDAR},
  pages={1516--1520},
  year={2019},
  url = {https://ieeexplore.ieee.org/abstract/document/8977955}
}

@article{fu2024ocrbench,
  title={Ocrbench v2: An improved benchmark for evaluating large multimodal models on visual text localization and reasoning},
  author={Fu, Ling and Kuang, Zhebin and Song, Jiajun and Huang, Mingxin and Yang, Biao and Li, Yuzhe and Zhu, Linghao and Luo, Qidi and Wang, Xinyu and Lu, Hao and others},
  journal={ArXiv preprint},
  year={2025},
  volume = {abs/2501.00321},
  url = {https://arxiv.org/abs/2501.00321}
}

@inproceedings{wang2024docllm,
  title={Docllm: A layout-aware generative language model for multimodal document understanding},
  author={Wang, Dongsheng and Raman, Natraj and Sibue, Mathieu and Ma, Zhiqiang and Babkin, Petr and Kaur, Simerjot and Pei, Yulong and Nourbakhsh, Armineh and Liu, Xiaomo},
  booktitle={Proceedings of ACL},
  pages={8529--8548},
  year={2024},
  url = {https://aclanthology.org/2024.acl-long.463/}
}

@inproceedings{yu2021pick,
  title={Pick: processing key information extraction from documents using improved graph learning-convolutional networks},
  author={Yu, Wenwen and Lu, Ning and Qi, Xianbiao and Gong, Ping and Xiao, Rong},
  booktitle={Proceedings of ICPR},
  url ={https://ieeexplore.ieee.org/document/9412927} ,
  pages={4363--4370},
  year={2021},
}

@article{cheng2022trie++,
  title={TRIE++: towards end-to-end information extraction from visually rich documents},
  author={Cheng, Zhanzhan and Zhang, Peng and Li, Can and Liang, Qiao and Xu, Yunlu and Li, Pengfei and Pu, Shiliang and Niu, Yi and Wu, Fei},
  journal={ArXiv preprint},
  year={2022},
  volume = {abs/2207.06744},
  url = {https://arxiv.org/abs/2207.06744}
}

@article{liu2024see,
  title={See then Tell: Enhancing Key Information Extraction with Vision Grounding},
  author={Liu, Shuhang and Zhang, Zhenrong and Hu, Pengfei and Ma, Jiefeng and Du, Jun and Wang, Qing and Zhang, Jianshu and Liu, Chenyu},
  journal={ArXiv preprint},
  year={2024},
  url = {https://arxiv.org/abs/2409.19573},
  volume = {abs/2409.19573},
}

@inproceedings{dhouib2023docparser,
  title={Docparser: End-to-end ocr-free information extraction from visually rich documents},
  author={Dhouib, Mohamed and Bettaieb, Ghassen and Shabou, Aymen},
  booktitle={Proceedings of ICDAR},
  pages={155--172},
  year={2023},
  url = {https://link.springer.com/chapter/10.1007/978-3-031-41734-4_10}
}

@article{lee2022formnet,
  title={Formnet: Structural encoding beyond sequential modeling in form document information extraction},
  author={Lee, Chen-Yu and Li, Chun-Liang and Dozat, Timothy and Perot, Vincent and Su, Guolong and Hua, Nan and Ainslie, Joshua and Wang, Renshen and Fujii, Yasuhisa and Pfister, Tomas},
  journal={ArXiv preprint},
  year={2022},
  url = {https://arxiv.org/abs/2203.08411},
  volume = {abs/2203.08411},
}

@article{hui2025seg,
  title={SEG-Doc: A simple yet efficient graph neural network framework for document key information extraction},
  author={Hui, Yangyang and Liu, Jiahao and Zhang, Qi and Zhou, Tianyi and Song, Yonghong},
  journal={Neurocomputing},
  pages={130493},
  year={2025},
  publisher={Elsevier},
  url = {https://www.sciencedirect.com/science/article/abs/pii/S0925231225011658}
}

@inproceedings{palm2017cloudscan,
  title={Cloudscan-a configuration-free invoice analysis system using recurrent neural networks},
  author={Palm, Rasmus Berg and Winther, Ole and Laws, Florian},
  booktitle={Proceedings of ICDAR},
  pages={406--413},
  year={2017},
  url = {https://ieeexplore.ieee.org/abstract/document/8270005}
}

@inproceedings{kim2022ocr,
  title={Ocr-free document understanding transformer},
  author={Kim, Geewook and Hong, Teakgyu and Yim, Moonbin and Nam, JeongYeon and Park, Jinyoung and Yim, Jinyeong and Hwang, Wonseok and Yun, Sangdoo and Han, Dongyoon and Park, Seunghyun},
  booktitle={Proceedings of ECCV},
  pages={498--517},
  year={2022},
  url = {https://link.springer.com/chapter/10.1007/978-3-031-19815-1_29}
}

@inproceedings{gao2011structure,
  title={Structure extraction from PDF-based book documents},
  author={Gao, Liangcai and Tang, Zhi and Lin, Xiaofan and Liu, Ying and Qiu, Ruiheng and Wang, Yongtao},
  booktitle={Proceedings of the 11th annual international ACM/IEEE joint conference on Digital libraries},
  pages={11--20},
  year={2011},
  url = {https://dl.acm.org/doi/abs/10.1145/1998076.1998079}
}

@inproceedings{skalicky2022business,
  title={Business document information extraction: Towards practical benchmarks},
  author={Skalick{\`y}, Maty{\'a}{\v{s}} and {\v{S}}imsa, {\v{S}}t{\v{e}}p{\'a}n and U{\v{r}}i{\v{c}}{\'a}{\v{r}}, Michal and {\v{S}}ulc, Milan},
  booktitle={International Conference of the Cross-Language Evaluation Forum for European Languages},
  pages={105--117},
  year={2022},
  organization={Springer},
  url = {https://ieeexplore.ieee.org/abstract/document/273022}
}

@inproceedings{yang2025cc,
  title={Cc-ocr: A comprehensive and challenging ocr benchmark for evaluating large multimodal models in literacy},
  author={Yang, Zhibo and Tang, Jun and Li, Zhaohai and Wang, Pengfei and Wan, Jianqiang and Zhong, Humen and Liu, Xuejing and Yang, Mingkun and Wang, Peng and Bai, Shuai and others},
  booktitle={Proceedings of ICCV},
  pages={21744--21754},
  year={2025},
  url = {https://openaccess.thecvf.com/content/ICCV2025/html/Yang_CC-OCR_A_Comprehensive_and_Challenging_OCR_Benchmark_for_Evaluating_Large_ICCV_2025_paper.html}
}

@article{yu2023icdar2023competitionstructured,
      title={ICDAR 2023 Competition on Structured Text Extraction from Visually-Rich Document Images}, 
      author={Wenwen Yu and Chengquan Zhang and Haoyu Cao and Wei Hua and Bohan Li and Huang Chen and Mingyu Liu and Mingrui Chen and Jianfeng Kuang and Mengjun Cheng and Yuning Du and Shikun Feng and Xiaoguang Hu and Pengyuan Lyu and Kun Yao and Yuechen Yu and Yuliang Liu and Wanxiang Che and Errui Ding and Cheng-Lin Liu and Jiebo Luo and Shuicheng Yan and Min Zhang and Dimosthenis Karatzas and Xing Sun and Jingdong Wang and Xiang Bai},
      year={2023},
      url = {https://arxiv.org/abs/2306.03287},
       volume = {abs/2306.03287},
      journal = {ArXiv preprint},
}

@inproceedings{yang2023modeling,
  title={Modeling Entities as Semantic Points for Visual Information Extraction in the Wild},
  author={Yang, Zhibo and Long, Rujiao and Wang, Pengfei and Song, Sibo and Zhong, Humen and Cheng, Wenqing and Bai, Xiang and Yao, Cong},
  url = {https://openaccess.thecvf.com/content/CVPR2023/html/Yang_Modeling_Entities_As_Semantic_Points_for_Visual_Information_Extraction_in_CVPR_2023_paper.html},
  booktitle={Proceedings of ICCV},
  year={2023}
}

@INPROCEEDINGS{8892998,
  author={Jaume, Guillaume and Kemal Ekenel, Hazim and Thiran, Jean-Philippe},
  booktitle={2019 International Conference on Document Analysis and Recognition Workshops (ICDARW)}, 
  title={FUNSD: A Dataset for Form Understanding in Noisy Scanned Documents}, 
  year={2019},
  pages={1-6},
  url ={https://ieeexplore.ieee.org/abstract/document/8892998}
}

@inproceedings{wang2021vies,
  title={Towards Robust Visual Information Extraction in Real World: New Dataset and Novel Solution},
  author={Wang, Jiapeng and Liu, Chongyu and Jin, Lianwen and Tang, Guozhi and Zhang, Jiaxin and Zhang, Shuaitao and Wang, Qianying and Wu, Yaqiang and Cai, Mingxiang},
  booktitle={Proceedings of AAAI},
  year={2021},
  url = {https://ojs.aaai.org/index.php/AAAI/article/view/16378}
}

\clearpage
\newpage
\appendix

\section{Appendix}
\label{sec:appendix}

\begin{table*}[!t]
\centering
\small
\begin{tabular}{l l c l c c}
\toprule
\textbf{Domain} & \textbf{Scenario} & \textbf{Abbreviation} &
\textbf{Data Source} & \textbf{\#Test Samples} & \textbf{Ratio} \\
\midrule
\multirow{4}{*}{Business Transactions}
& Commercial        & Com.  & SIBR, DocILE & 620 & 13.86\% \\
& Retail            & Ret.  & SROIE        & 347 & 7.76\%  \\
& Catering Services & Cat.  & CORD, CELL   & 212 & 4.74\%  \\
& Accommodation     & Acco. & SIBR         & 40  & 0.89\%  \\
\midrule
\multirow{4}{*}{Public Services}
& Administrative    & Admi. & CELL, FUNSD  & 385 & 8.61\%  \\
& Education         & Edu.  & EPHOIE, CELL & 320 & 7.16\%  \\
& Postal Label    & Post. & HW-FORMS     & 500 & 11.18\% \\
& Advertisement         & Adv.  & DeepForm     & 71  & 1.59\%  \\
\midrule
\multirow{3}{*}{Regulated Records}
& Tax-Compliant     & Tax.  & Nanonets-KIE & 987 & 22.07\% \\
& Medical Services  & Med.  & SIBR         & 240 & 5.37\%  \\
& Nutrition Label   & Nut.  & POIE         & 750 & 16.77\%\\
\bottomrule
\end{tabular}
\caption{Dataset Sources for the Constrained-Category KIE Track of \method{}.}
\label{tab:source_for_cc_track}
\end{table*}
\subsection{License}
We strictly comply with the original licenses released with each dataset used to build \method{} and do not redistribute any third-party raw document images.  Among the datasets used in this work, DocILE and SROIE are released under the MIT License; CORD and SIBR are distributed under the CC BY-SA 4.0 License; and Nanonets-KIE is provided under the Apache License 2.0. In addition, FUNSD, EPHOIE, HW-FORMS, and CELL are restricted to non-commercial academic use and are included in this study in accordance with their original licensing terms. DeepForm is subject to the ProPublica Terms of Use\footnote{\url{https://projects.propublica.org/datastore/terms}}. For datasets without an explicitly stated standard license, we follow the original release terms, use them solely for research evaluation, and do not redistribute the raw data.

\subsection{Data Sources for Documents in the Constrained-Category KIE Track}
\label{app:datac}

The document images in the constrained-category KIE track are collected from multiple existing public document understanding datasets. We detail the data sources in Table~\ref{tab:source_for_cc_track}. Specifically, we curate documents across 3 domains: Business Transactions, Public Services, and Regulated Records, and further organize them into 11 representative application scenarios. 

For Business Transactions, we include documents from commercial, retail, catering, and accommodation scenarios. These documents are mainly drawn from SIBR~\cite{yang2023modeling}, DocILE~\cite{vsimsa2023docile}, SROIE~\cite{huang2019icdar2019}, CORD~\cite{park2019cord}, and CELL~\cite{yu2023icdar2023competitionstructured}, covering typical transactional records such as receipts, invoices, and service-related documents. The Public Services domain focuses on administrative and societal documents, including administrative forms, educational documents, postal labels, and advertisements. Data in this domain are collected from CELL~\cite{yu2023icdar2023competitionstructured}, FUNSD~\cite{8892998}, EPHOIE~\cite{wang2021vies}, HW-FORMS\footnote{\url{https://huggingface.co/datasets/ift/handwriting_forms}}, and DeepForm\footnote{\url{https://github.com/jstray/deepform}}. Compared to business documents, these samples exhibit greater layout diversity and more heterogeneous information requirements. The Regulated Records domain consists of documents subject to stricter formatting or compliance requirements, including tax-compliant documents, medical service records, and nutrition labels. We filter these documents from Nanonets-KIE\footnote{\url{https://huggingface.co/datasets/nanonets/key_information_extraction}}, SIBR~\cite{yang2023modeling}, and POIE~\cite{kuang2023visual}. This domain typically involves complex schemas, fine-grained field definitions, and higher sensitivity to extraction errors, making it particularly challenging for end-to-end KIE systems.

\subsection{Annotation Details for the Constrained-Category KIE Track}
\label{app:ann}
To curate annotations for the constrained-category KIE track, we adopt a systematic annotation curation pipeline. Specifically, we distinguish between datasets with existing KIE annotations and those that provide only OCR annotations, and apply specific processing strategies accordingly.

For datasets with existing KIE annotations, we directly use the raw field annotations or map them to our schema. However, the quality of these annotations varies substantially across datasets, with common issues including annotated field values that do not appear in the document, inconsistent capitalization, overlaps with irrelevant text, and incomplete or partially missing values. To address these issues, we correct misannotated field values, remove overlaps with irrelevant text regions, normalize formatting inconsistencies, and supplement missing values when they can be unambiguously inferred from the document. Annotations that remain ambiguous or unsupported are excluded to ensure high annotation reliability. For datasets that only provide OCR annotations that can not be converted to the end-to-end KIE formulation, we manually re-annotate them based on the scenario-predefined schema we identified before. 

\subsection{Data Curation Details for the Open-Category KIE Track}
\label{app:open_curation}
To construct the open-category KIE track in our benchmark, we utilize a document reconstruction pipeline to collect documents, ensuring the diversity and realism of document content and layouts while avoiding privacy and copyright concerns inherent in real-world documents.

For each language and document type, we curate representative real-world documents from practical scenarios and use them as in-context demonstrations, which are selected to reflect the typical content elements of the target document type. These documents are not reused in the benchmark and are solely intended to anchor the generation in realistic document patterns. 
Then, we prompt GPT-4o to generate a description for the target document type based on these demonstrations. These descriptions characterize the types of content elements and their structural organization (\textit{e.g.}, headers, keys, tables, and free-form text blocks), without specifying concrete field values or copying real-world content, preventing the leakage of sensitive or proprietary information. Based on the description, we further prompt the model to instantiate a concrete document by generating executable HTML that explicitly specifies both the textual elements and their layout. To enhance the diversity of the documents, we follow~\citet{ge2024scaling} and assign a distinct persona to each sample, conditioning generation on persona-specific attributes so that the resulting documents exhibit richer variation in content and presentation. We then render the HTML code into document images and remove samples with obvious rendering failures, obtaining the document images used in the open-category KIE track.

We further introduce lightweight noise into the rendered document images to better approximate the imperfections commonly observed in real-world scanned or photographed documents. Specifically, we first model the document images as 3D objects in Blender\footnote{\url{https://www.blender.org}} to simulate diverse lighting conditions and crease effects encountered in real-world scenarios. This 3D-based rendering process closely mirrors practical document acquisition settings, enabling synthesized images to capture realistic illumination variations, shading patterns, and surface deformations that are difficult to reproduce with purely 2D augmentations. This yields visual characteristics that are nearly indistinguishable from those observed in real scanned or photographed documents. We then capture the document images using a vertically positioned camera and apply a diverse set of perturbations that simulate common acquisition and printing artifacts, including motion blur and mild Gaussian blur to mimic camera shake or defocus, elastic distortions to emulate paper deformation during scanning, and slight perspective transformations to reflect imperfect capture angles. Additionally, we introduce printer-related artifacts, including thin ink streaks, faded or broken ink regions, subtle variations in ink color, and horizontal noise patterns that resemble dirty drum effects. These perturbations are applied with low intensity and carefully constrained so as not to alter textual content or layout. Samples in which noise perturbations cause excessive degradation to textual content are manually identified and removed, ensuring that all retained documents remain legible and semantically reliable.

\subsection{Detailed Analysis of the Documents in the Open-Category KIE Track}
\label{app:open_analysis}

The open-category KIE track is constructed from automatically generated document images, which naturally raises questions about whether these documents adequately capture the realism and diversity of real-world inputs encountered in practical KIE scenarios. To address this concern, we conduct a comprehensive analysis from multiple complementary perspectives, examining document realism, content and field diversity, as well as layout variability.

\begin{figure}[t]
    \centering
    \begin{subfigure}[t]{0.462\linewidth}
        \centering
                \includegraphics[width=\linewidth]{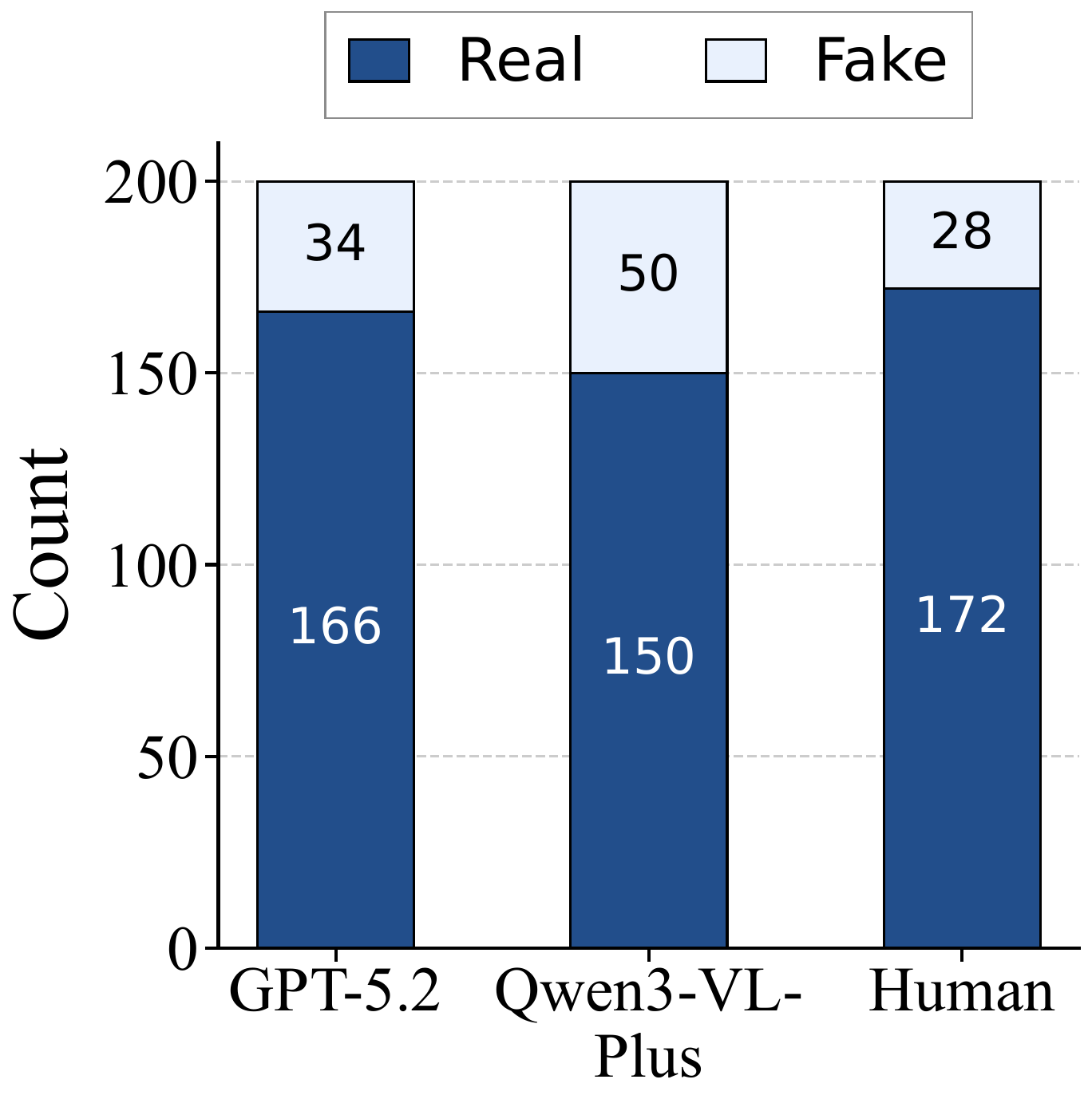}
        \caption{Document Authenticity Analysis by Human and Advanced LMMs.}
        \label{fig:human_and_lmms}
    \end{subfigure}
    \hfill
    \begin{subfigure}[t]{0.458\linewidth}
        \centering
        \includegraphics[width=\linewidth]{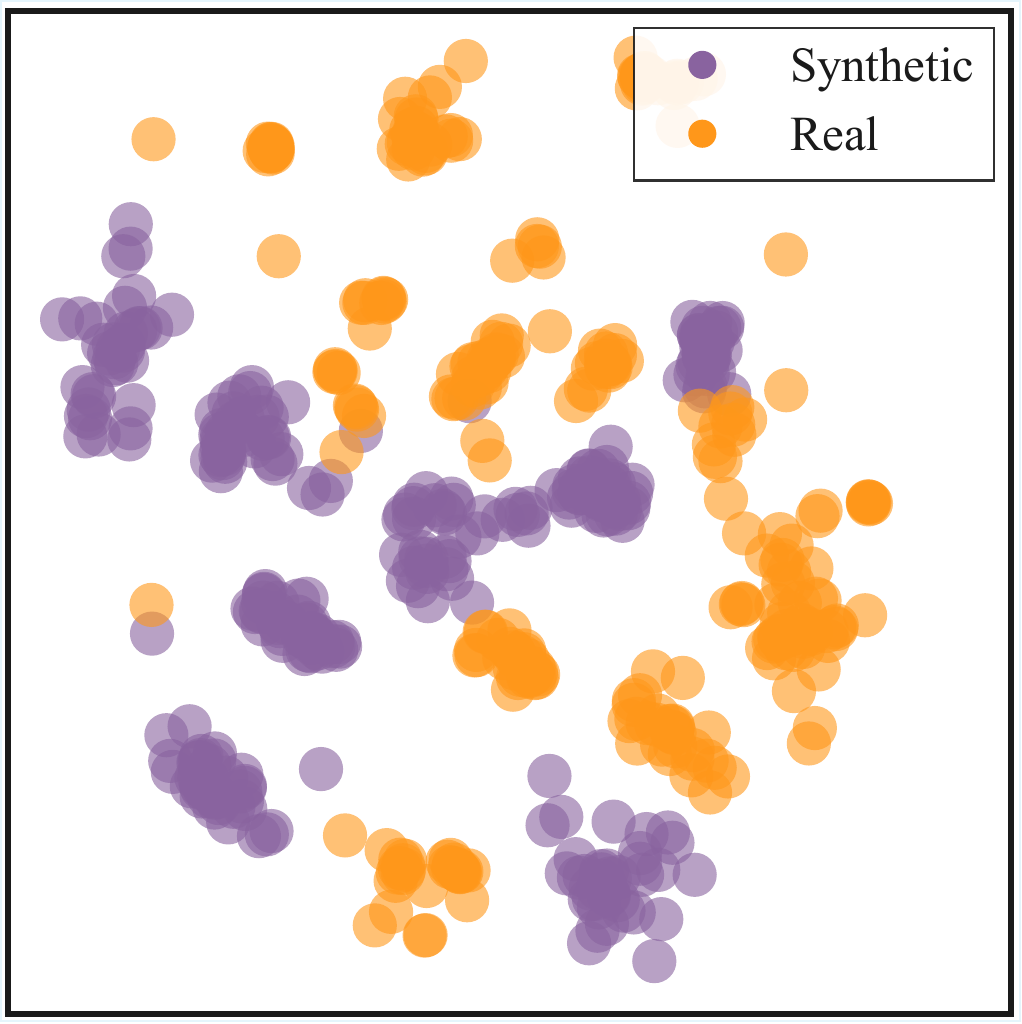}
        \caption{ Visualization of Document-Level Embeddings for Synthesized and Real Documents.}
        \label{fig:tsne_auth}
    \end{subfigure}
    \caption{Document Authenticity
Analysis in the Open-Category KIE Track.}
\end{figure}

\textbf{Authenticity Analysis.} 
Figure~\ref{fig:human_and_lmms} reports the authenticity judgments on the 200 synthesized document images in the open-category KIE track. Each evaluator is asked to determine whether a document is real or generated, without access to any auxiliary information. 
We find that only a limited number of synthesized documents are correctly identified as generated. Specifically, GPT-5.2 correctly classifies 34 out of the 200 synthesized documents as generated, while Qwen3-VL-Plus correctly identifies 50 synthesized documents. Human evaluators recognize 28 synthesized documents as generated. The majority of synthesized documents are instead judged as real, indicating that they are perceptually similar to real-world documents.

In addition to subjective authenticity judgments, we further investigate the distributional similarity between synthesized documents and real documents from the representation perspective, as shown in Figure~\ref{fig:tsne_auth}. Specifically, we employ a multimodal embedding model, gme-Qwen2-VL-2B-Instruct\footnote{\url{https://huggingface.co/Alibaba-NLP/gme-Qwen2-VL-2B-Instruct}}, to extract document embeddings and project both real and synthesized documents into a shared low-dimensional embedding space for visualization and comparison. The real documents are sampled from the constrained-category KIE track.
The results show that the two types of documents are largely intermingled, rather than forming clearly separable clusters. This observation indicates that the synthesized documents closely resemble real documents in terms of their global visual–semantic representations. These findings provide additional evidence that the generated documents achieve a strong distributional alignment with real-world documents.

\begin{figure}[t]
    \centering
    \begin{subfigure}[t]{0.48\linewidth}
        \centering
        \includegraphics[width=\linewidth]{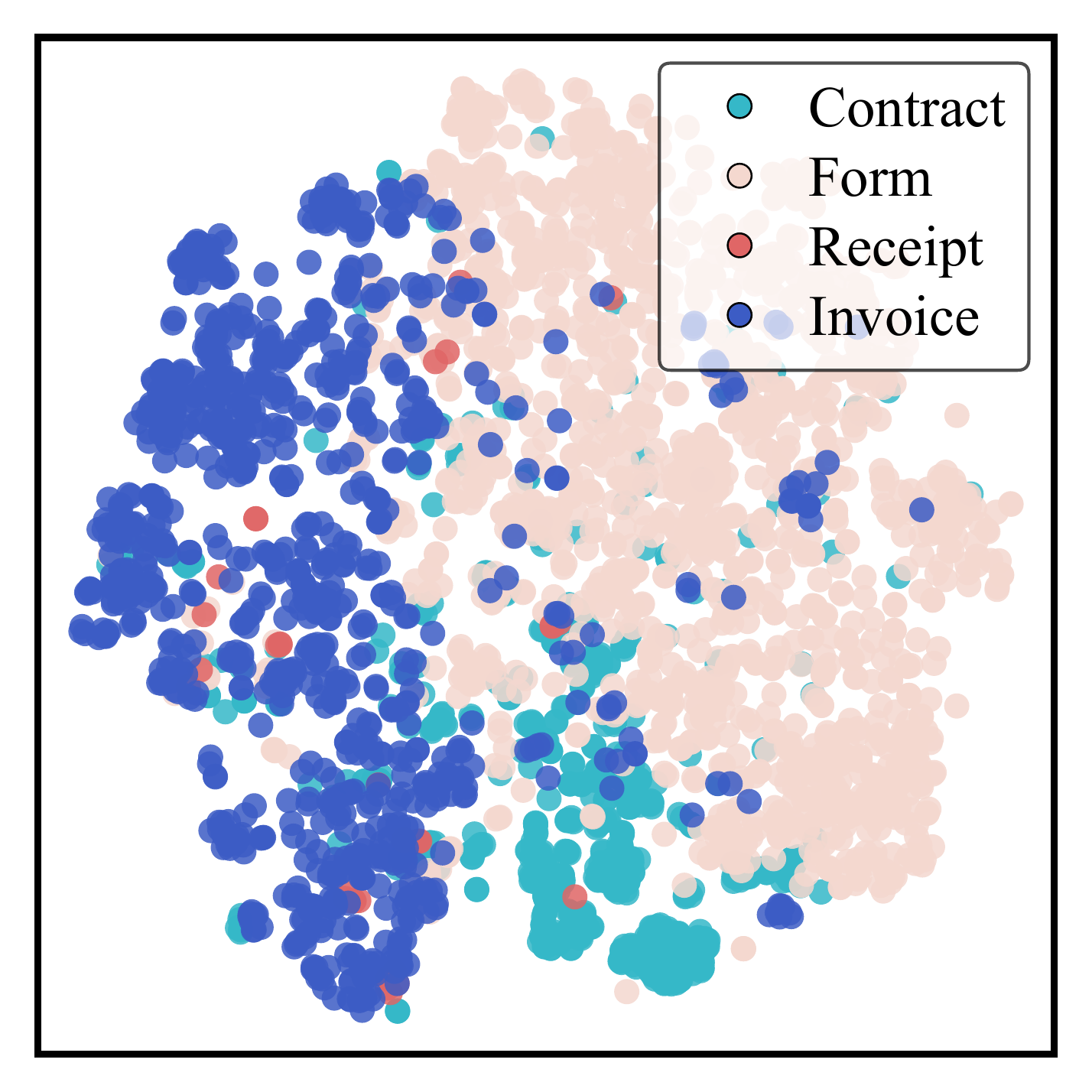}
    \caption{Visualization of Document Field Distribution in the Open-Category KIE Track.}
        \label{fig:tsne}
    \end{subfigure}
    \hfill
    \begin{subfigure}[t]{0.48\linewidth}
        \centering
        \includegraphics[width=\linewidth]{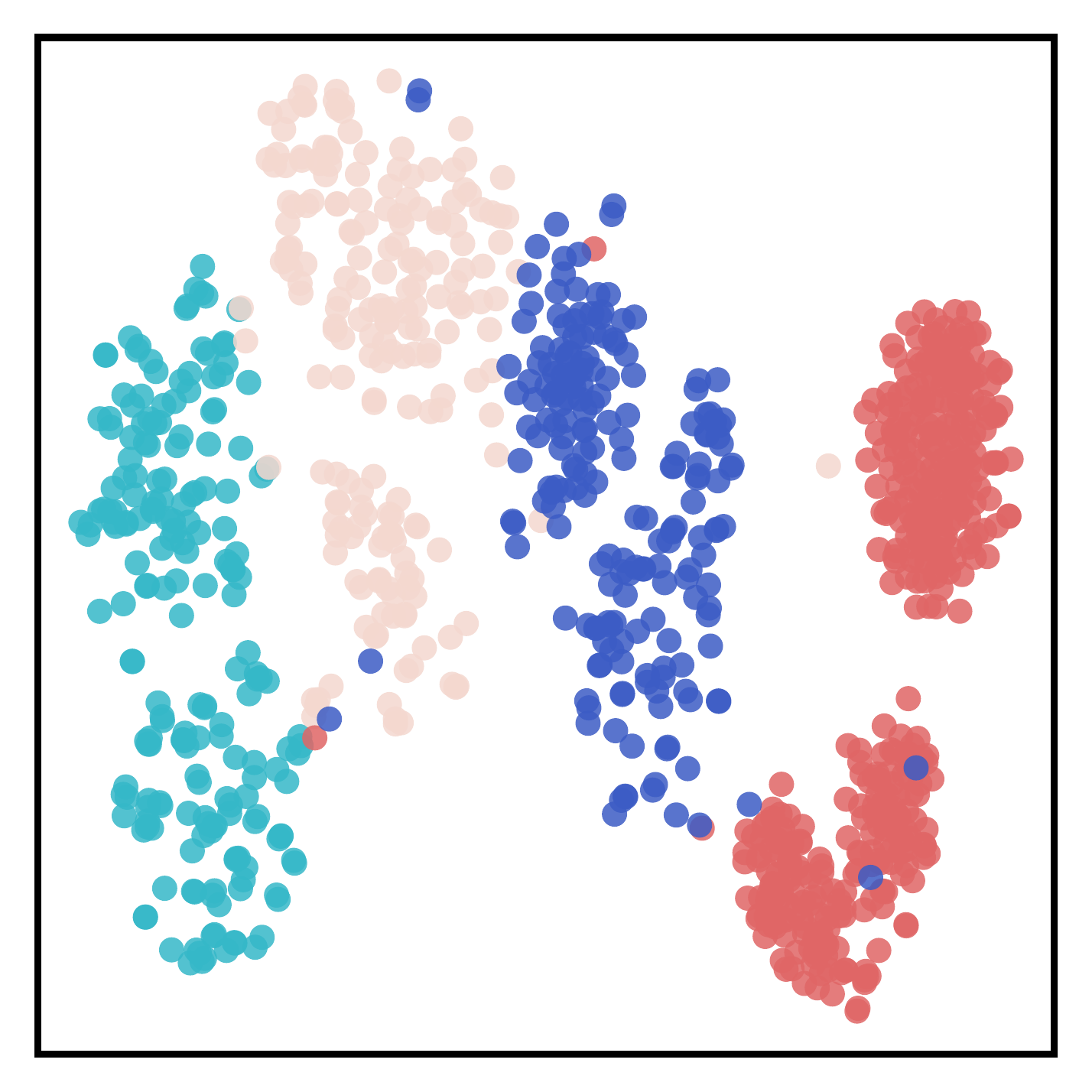}
        \caption{Visualization of Document Content Distribution in the Open-Category KIE Track.}
        \label{fig:content_tsne}
    \end{subfigure}
    \caption{Diversity Analysis of Document Images in the Open-Category KIE Track.}
    \label{fig:semantic_anal}
\end{figure}

\textbf{Semantic Analysis.} 
We further investigate whether the open-category KIE track exhibits sufficient semantic diversity, rather than degenerating into a narrow set of templated field schemas or repetitive content patterns. To this end, we analyze document semantics from two complementary perspectives—field-level distributions and content-level distributions—as illustrated in Figure~\ref{fig:semantic_anal}. In both analyses, we employ the Qwen3-Embedding model\footnote{\url{https://huggingface.co/Qwen/Qwen3-Embedding-8B}} to encode field values and full document contents into a shared semantic embedding space.

Figure~\ref{fig:tsne} visualizes the distribution of document fields in a shared semantic embedding space. We observe that fields associated with different document types are largely intermingled rather than forming isolated, document-type-specific clusters. This pattern suggests that the open-category KIE track does not rely on rigid field templates tied to individual document categories. Instead, similar field semantics naturally occur across heterogeneous document types, while documents of the same type can exhibit diverse field compositions. Such behavior aligns with real-world KIE scenarios, where field definitions are flexible and often context-dependent.

\begin{figure*}[t]
    \centering
    \begin{subfigure}[t]{0.32\linewidth}
        \centering
        \includegraphics[width=\linewidth]{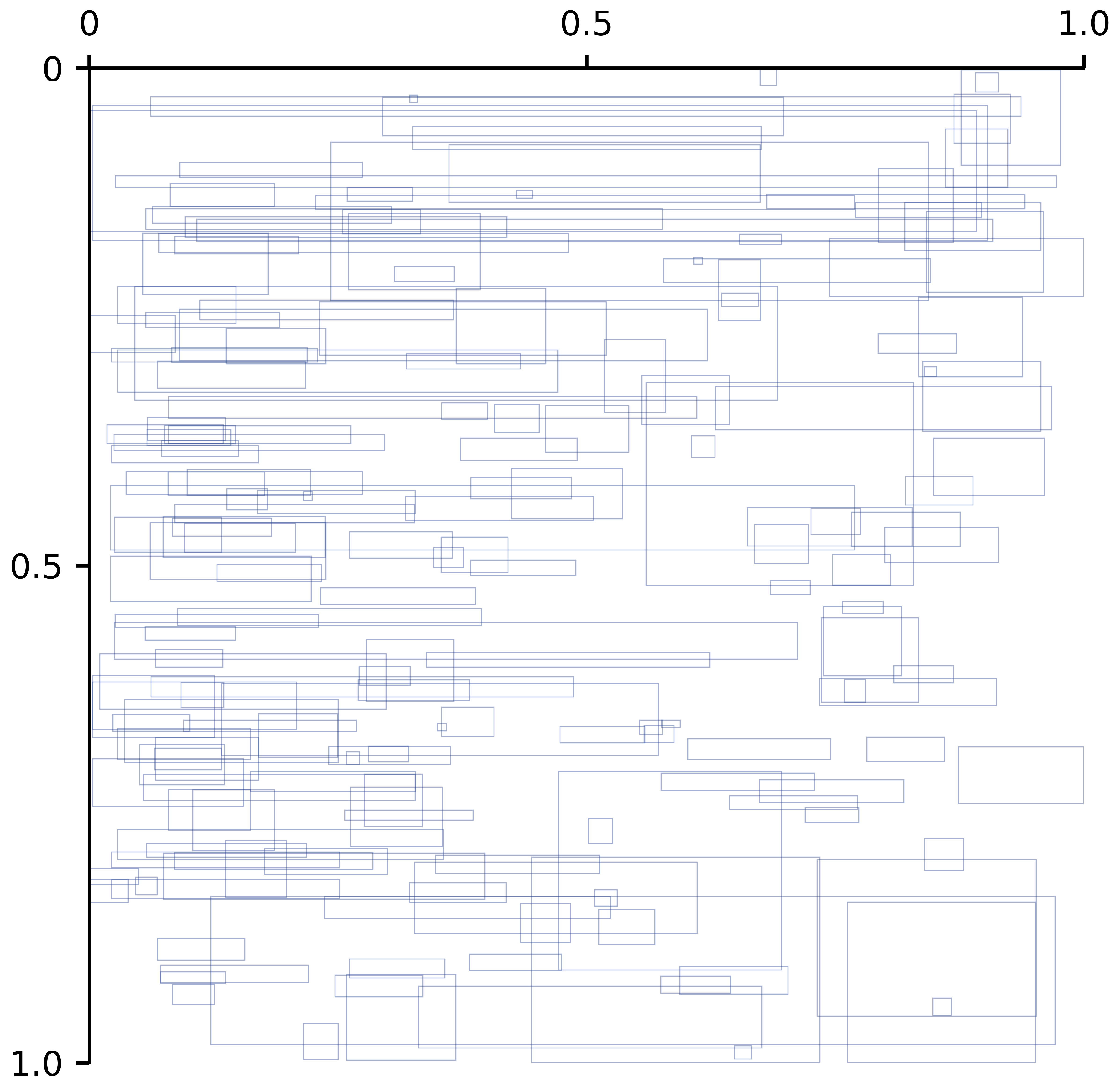}
        \caption{N=50}
        \label{fig:bbox_50}
    \end{subfigure}
    \hfill
    \begin{subfigure}[t]{0.32\linewidth}
        \centering
        \includegraphics[width=\linewidth]{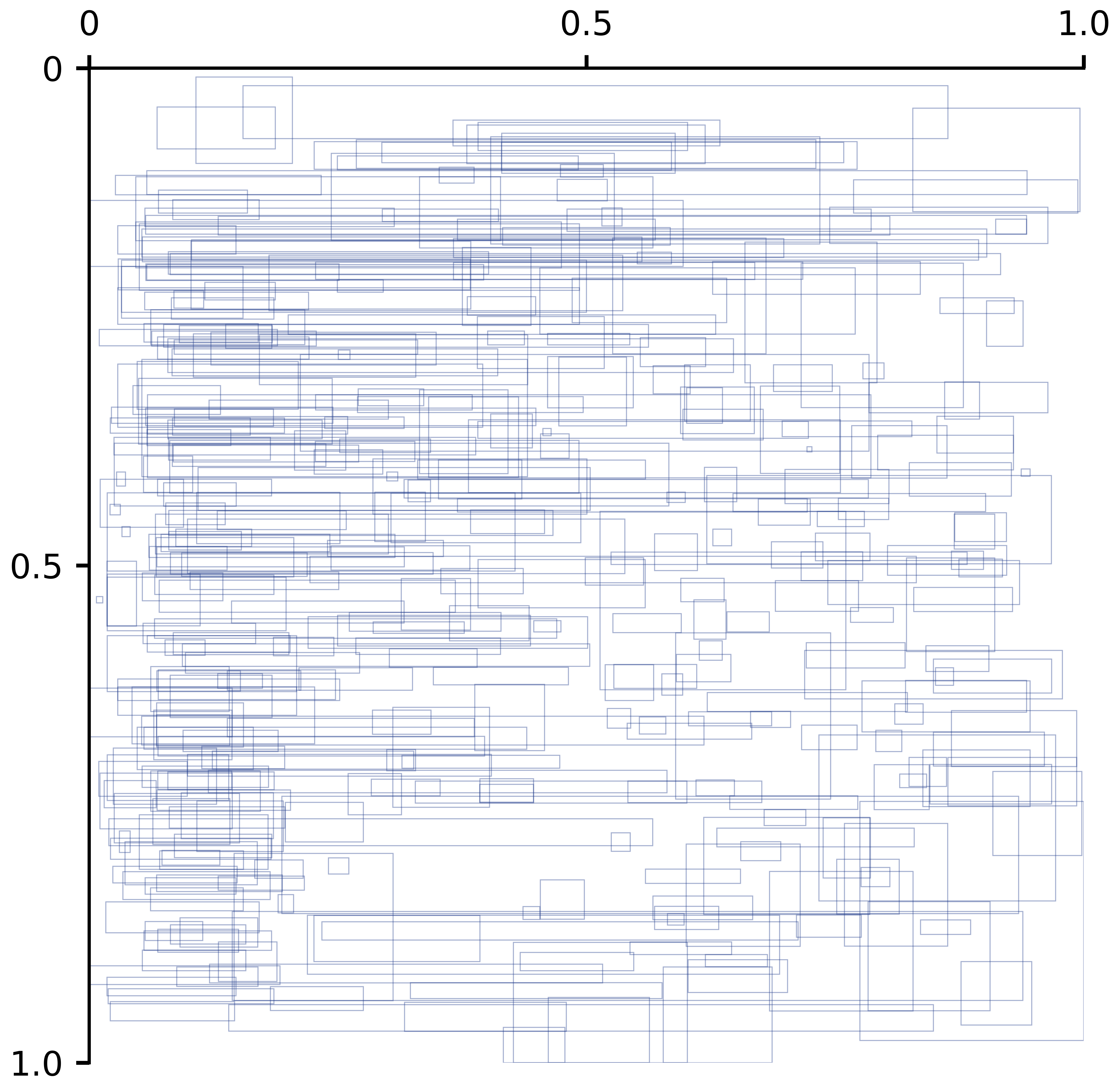}
        \caption{N=100}
        \label{fig:bbox_100}
    \end{subfigure}
    \hfill
    \begin{subfigure}[t]{0.32\linewidth}
        \centering
        \includegraphics[width=\linewidth]{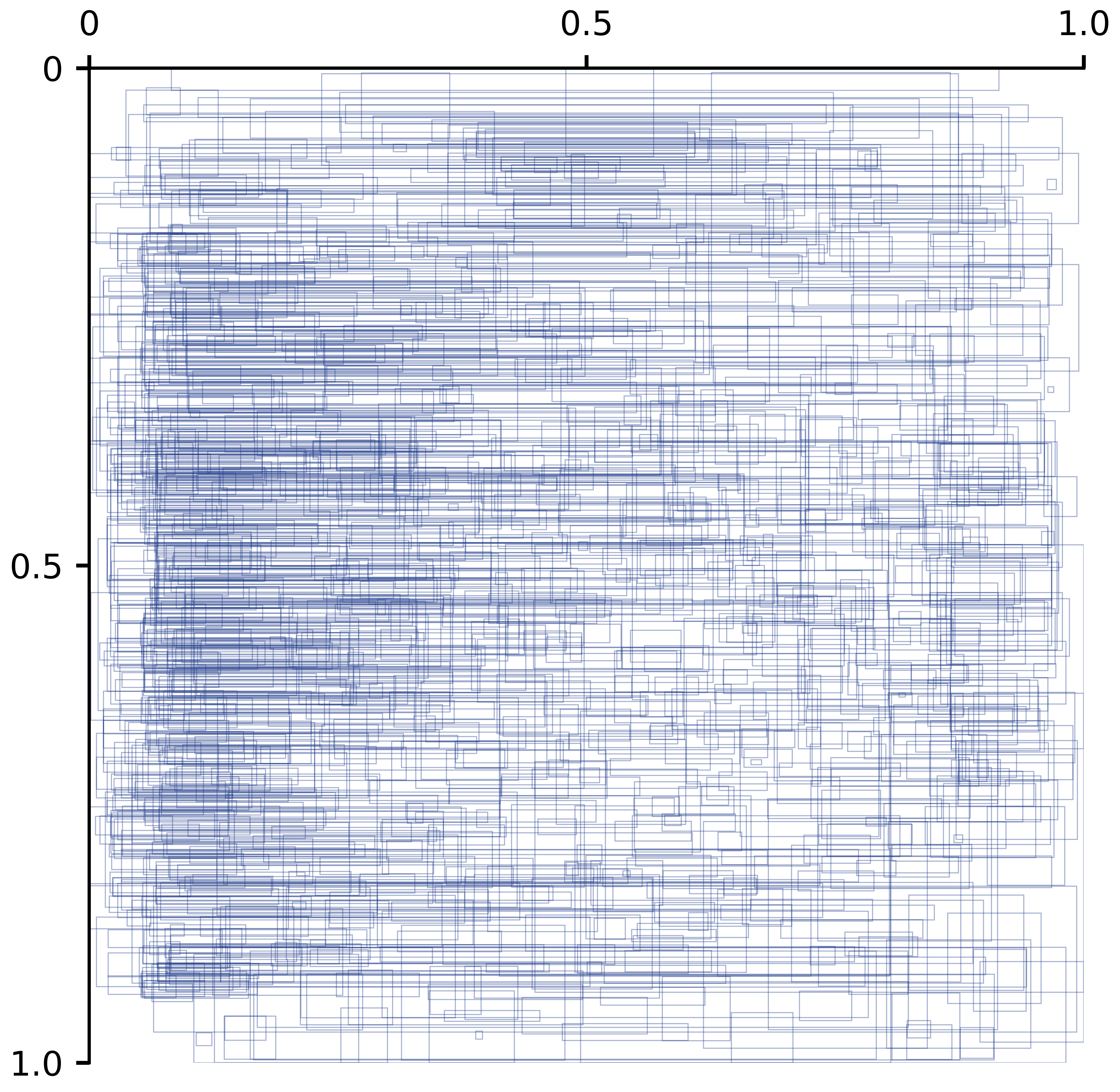}
        \caption{N=300}
        \label{fig:bbox_300}
    \end{subfigure}
    \caption{Visualization of Document Layout Diversity in the Open-Category KIE Track.}
    \label{fig:bbox_layout}
\end{figure*}

Figure~\ref{fig:content_tsne} presents the distribution of document content embeddings. In contrast to the field-level visualization, content representations exhibit clearer clustering across document types, reflecting semantic differences in document content. 
Each cluster shows substantial intra-class dispersion, indicating that the generated documents avoid trivial templating and instead cover a broad range of lexical and semantic variations within each category. This combination of inter-class separability and intra-class diversity suggests that the synthesized documents capture meaningful content diversity.

\textbf{Layout Analysis.}
We analyze layout diversity by visualizing where text regions appear on the document page. Figure~\ref{fig:bbox_layout} overlays normalized bounding boxes of text regions sampled from the synthesized documents into a shared coordinate space. Each subfigure corresponds to a different number of sampled bounding boxes $N$, allowing us to progressively reveal the underlying layout distribution as more text regions are included.

When $N$ is small, the visualization already exhibits multiple horizontal text lines and block-level regions distributed across different positions. As $N$ increases, these patterns become more pronounced: dense horizontal bands emerge, reflecting line-based text layouts commonly observed in real documents, while the wide spread in bounding box locations and sizes indicates substantial variation in layout structure. Importantly, even with a large number of bounding boxes overlaid, the layout does not collapse into a single dominant template.

\subsection{The Prompt Template Used in \method{}}
\label{app:prompt}
\begin{figure*}
    \centering
    \includegraphics[width=\linewidth]{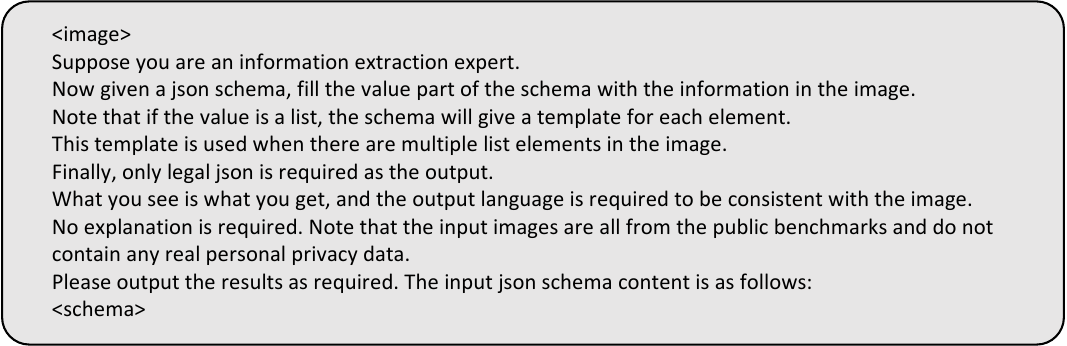}
    \caption{The Prompt Template for LMMs Used in \method{}.}
    \label{fig:prompt}
\end{figure*}

To ensure fairness and reproducibility, we use the same prompt template for all evaluated tracks and models as shown in Figure~\ref{fig:prompt}.
Specifically, we modify the prompt used by~\citet{yang2025cc} and remove explicit descriptions of the field, thereby requiring the model to infer the semantics of each field solely from its name and context. 

\begin{table*}[t]
\centering
\small
\begin{tabular}{l l p{0.45\linewidth} r}
\toprule
\textbf{Model} & \textbf{Vendor} & \textbf{Identifier} & \textbf{Release Date} \\
\midrule
\multicolumn{4}{l}{\cellcolor[rgb]{0.95,0.95,0.95}\text{Closed-source LMMs}} \\
\midrule
GPT-5 & OpenAI & gpt-5-2025-08-07 & 2025-08 \\
GPT-4o & OpenAI & gpt-4o-2024-08-16 & 2024-05 \\
Gemini-3-Pro & Google & gemini-3-pro-preview & 2025-11 \\
Claude-Sonnet-4.5 & Anthropic & claude-sonnet-4-5-20250929 & 2025-09 \\
Qwen3-VL-Plus & Alibaba & qwen3-vl-plus-2025-09-23 & 2025-09 \\
Qwen-VL-Max & Alibaba & qwen-vl-max-2025-08-13 & 2025-08 \\
\midrule
\multicolumn{4}{l}{\cellcolor[rgb]{0.95,0.95,0.95}\text{Open-source LMMs}} \\
\midrule
MiniCPM-V-4.5-8B & OpenBMB & OpenBMB/MiniCPM-V-4\_5 & 2025-08 \\
Gemma-3-12B & Google & google/gemma-3-12b-it & 2025-03 \\
InternVL3.5-8B & OpenGVLab & OpenGVLab/InternVL3\_5-8B-Instruct & 2025-08 \\
GLM-4.1V-9B & Zhipu AI & zai-org/GLM-4.1V-9B-Thinking & 2025-07 \\
Kimi-VL-A3B & Kimi & moonshotai/Kimi-VL-A3B-Instruct & 2025-04 \\
MiMo-VL-7B-RL & Xiaomi & XiaomiMiMo/MiMo-VL-7B-RL & 2025-06 \\
Qwen3-VL-8B & Alibaba & Qwen/Qwen3-VL-8B-Instruct & 2025-10 \\
SmolVLM2-2.2B & HuggingFace & HuggingFaceTB/SmolVLM2-2.2B-Instruct & 2025-02 \\
Ministral-3-8B & Mistral & mistralai/Ministral-3-8B-Instruct-2512 & 2025-12 \\
\bottomrule
\end{tabular}
\caption{Model Information of the Evaluated LMMs.}
\label{tab:model_details}
\end{table*}

\subsection{Detailed LMM Configurations}
\label{app:more}
In this section, we provide detailed information for the evaluated LMMs and report additional results on more models. Specifically, we show the model vendors, identifiers, and release dates in Table~\ref{tab:model_details}, ensuring transparency and reproducibility of our evaluation. For closed-source models, the identifiers correspond to the official names used in their SDKs, while for open-source models, they refer to the corresponding repositories on HuggingFace\footnote{\url{https://huggingface.co/}}.
All open-source models are deployed on two NVIDIA A100 GPUs.

\subsection{Behavioral Analysis for LMMs in KIE}
\label{app:behavior}
In this subsection, we conduct an in-depth investigation of LMM behaviors when performing the KIE task. Specifically, we take Qwen3-VL-8B-Instruct~\cite{bai2025qwen3vltechnicalreport} as a representative model and analyze its attention patterns layer by layer, aiming to elucidate the behavioral mechanisms underlying the extraction process.

\begin{figure*}
    \centering
    \includegraphics[width=\linewidth]{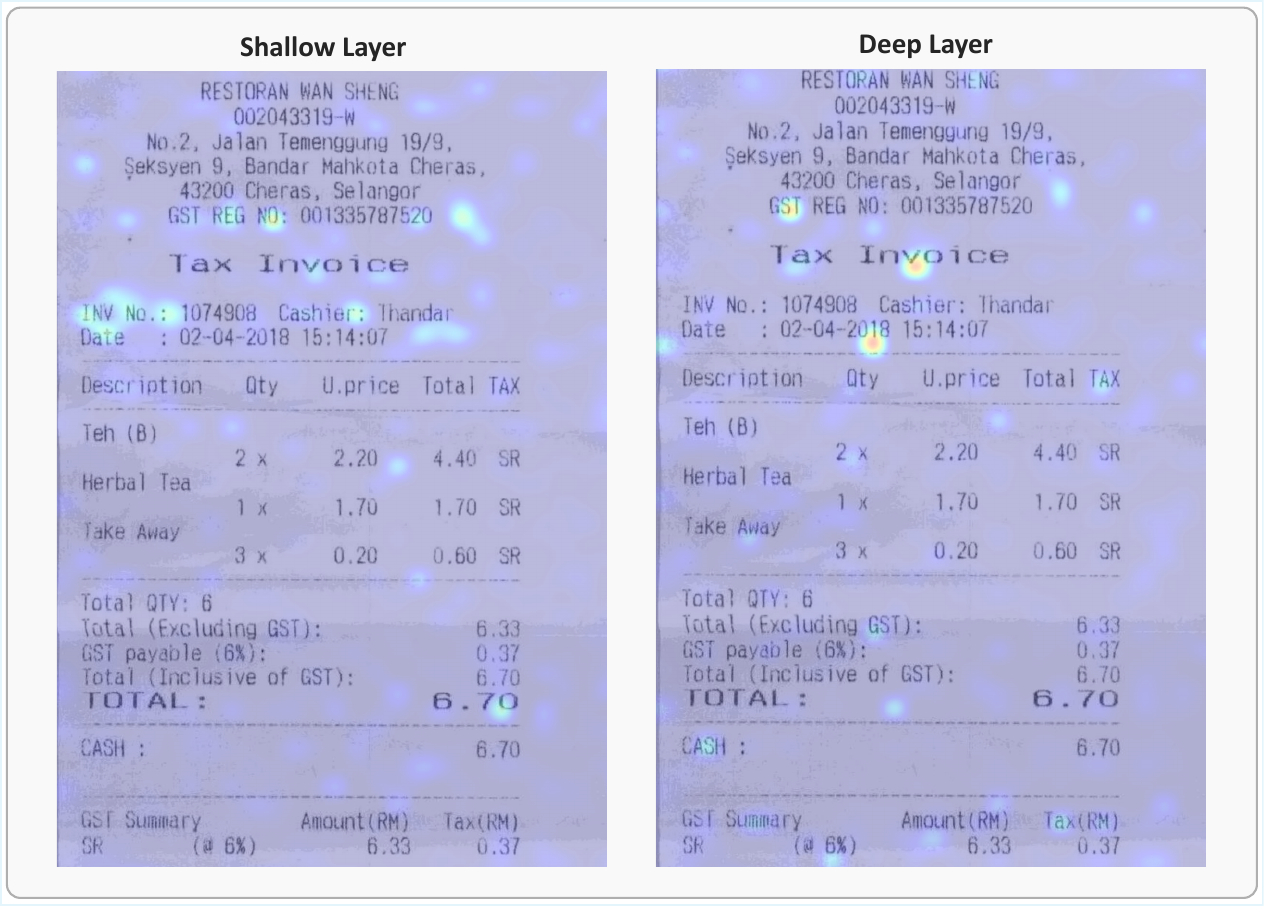}
    \caption{Layer-wise Attention Transition with Explicit Field Cues.}
    \label{fig:atten_case1}
\end{figure*}

In the first scenario, as illustrated in Figure~\ref{fig:atten_case1}, where the field name is explicitly visible in the document, the model demonstrates a clear layer-wise transition in its attention patterns. 
In the shallow layers, attention is predominantly concentrated on the field label itself, suggesting that the model initially searches for and anchors to explicit field cues present in the document.
As processing progresses into deeper layers, the model’s attention gradually shifts away from the field label toward the corresponding field value. This transition indicates that the model first establishes a semantic anchor via the field cue and then leverages this anchor to locate and extract the associated value, thereby completing the extraction process in a relatively direct and localized manner.

\begin{figure*}
    \centering
    \includegraphics[width=\linewidth]{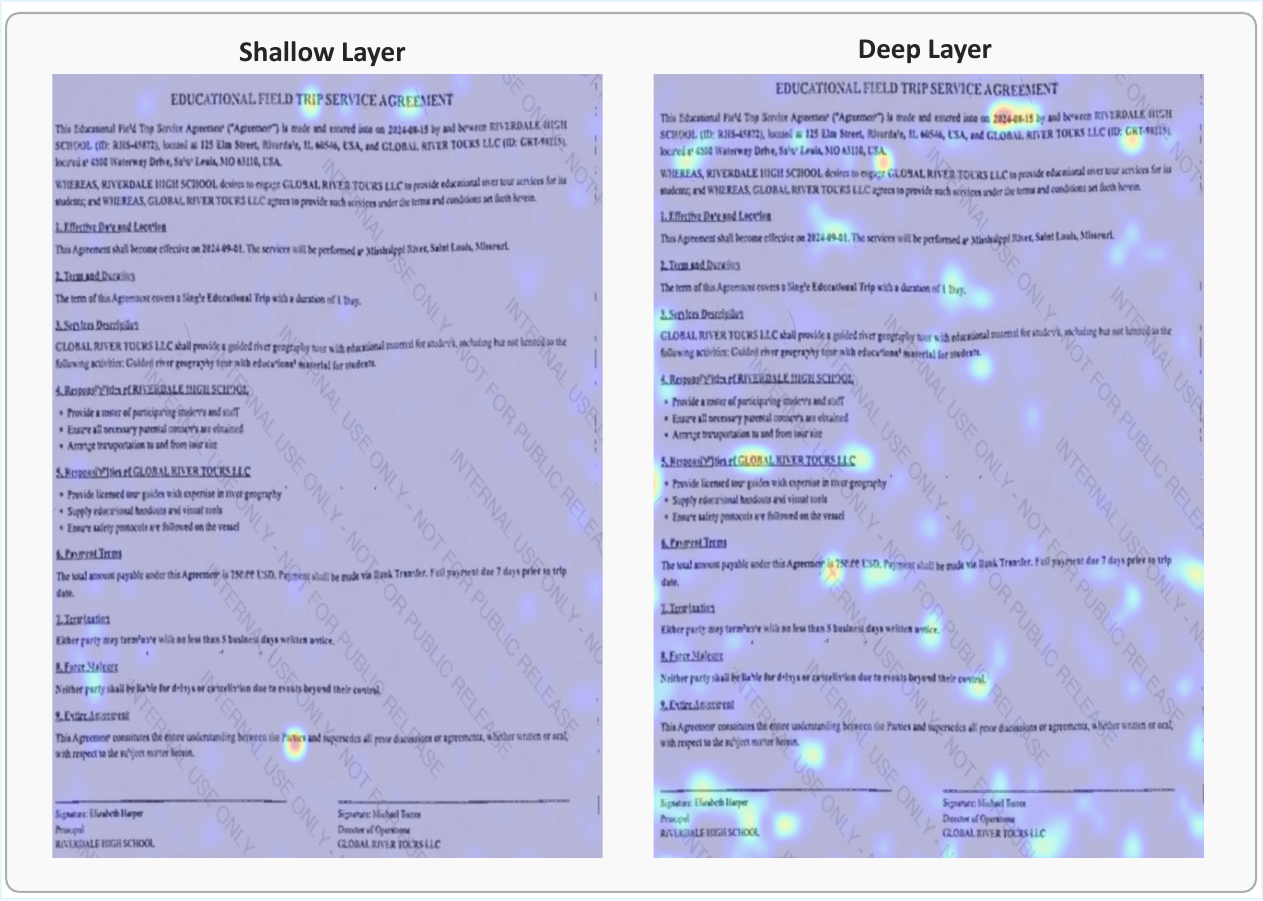}
    \caption{Layer-wise Attention  Transition under Implicit Field Cues.}
    \label{fig:atten_case2}
\end{figure*}

In contrast, the second scenario corresponds to cases where the field name does not explicitly appear, and the target value is implicitly embedded within a broader textual context, as illustrated in the Figure~\ref{fig:atten_case2}.
In this scenario, shallow-layer attention is predominantly allocated to approximate or semantically related field-like phrases, suggesting that the model initiates the extraction process by hypothesizing and probing potential semantic anchors in the absence of explicit field cues. 
At deeper layers, the model’s attention no longer converges on a single location; instead, it is distributed across multiple candidate text spans that potentially contain the target value. This behavior suggests that the model engages in implicit candidate aggregation and comparison, inferring the correct extraction primarily through contextual reasoning rather than relying on explicit field cues.

\subsection{Examples from \method{}}
To offer a more concrete perspective on the challenges posed by \method{}, we present several representative document examples in this subsection.
These examples span multiple evaluation tracks, document types, and application scenarios, collectively demonstrating the extensive visual, structural, and semantic diversity captured by the benchmark.

\begin{figure*}
    \centering
    \includegraphics[width=\linewidth]{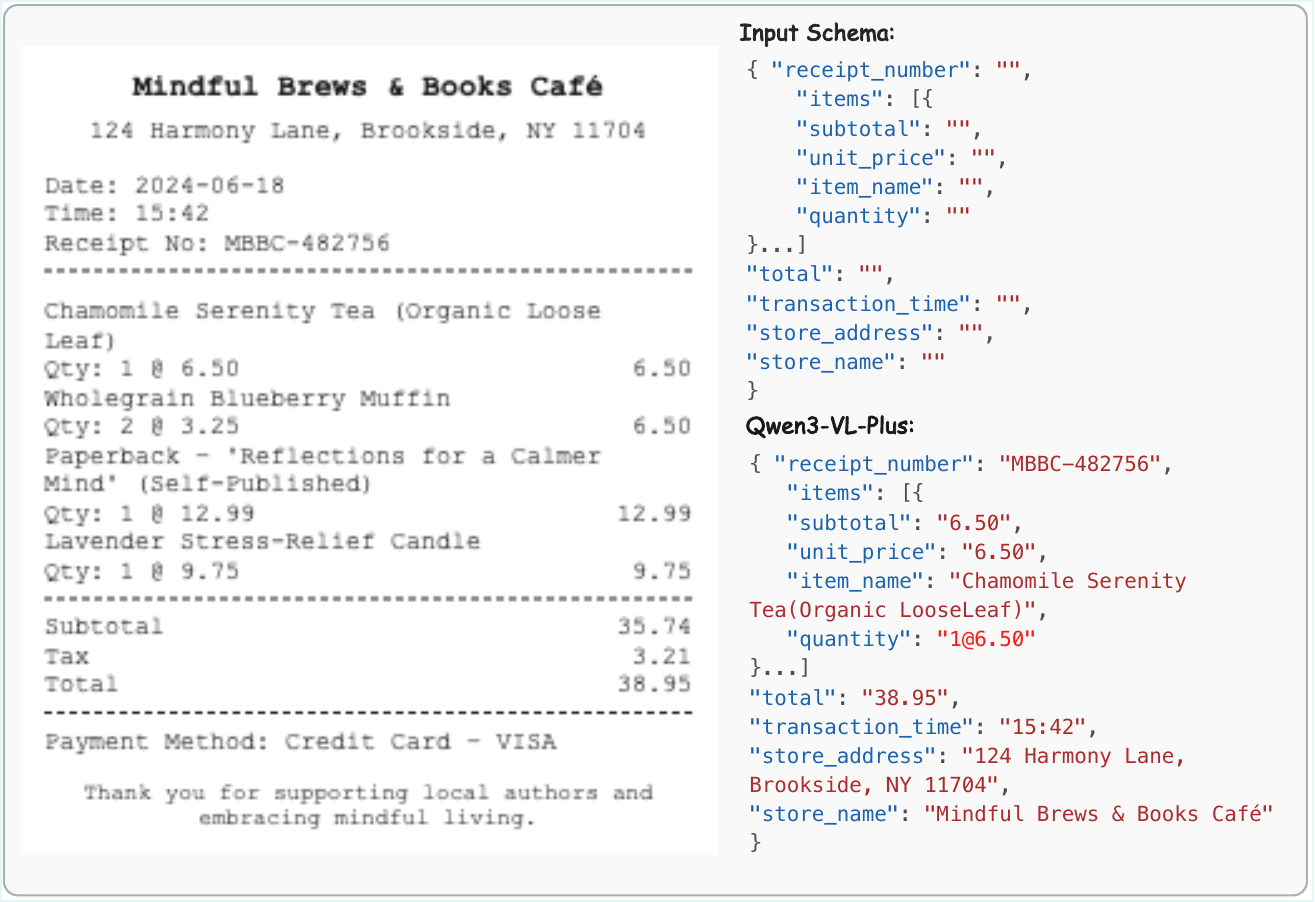}
    \caption{Example from the Receipt Document Type in the Open-Category KIE Track.}
    \label{fig:case1}
\end{figure*}

Figure~\ref{fig:case1} shows a representative receipt example from the open-category KIE track. The document exhibits a typical real-world receipt layout, with dense text, multiple itemized entries, and both explicit structural cues and implicit information embedded in free-form text. Given the input schema, the model is required to extract receipt-level metadata as well as variable-length item records. As illustrated by the prediction, even strong LMMs may produce subtle field interpretation errors, such as confusing quantities with unit prices, highlighting the difficulty of robust structured extraction under realistic layouts.

\begin{figure*}
    \centering
    \includegraphics[width=\linewidth]{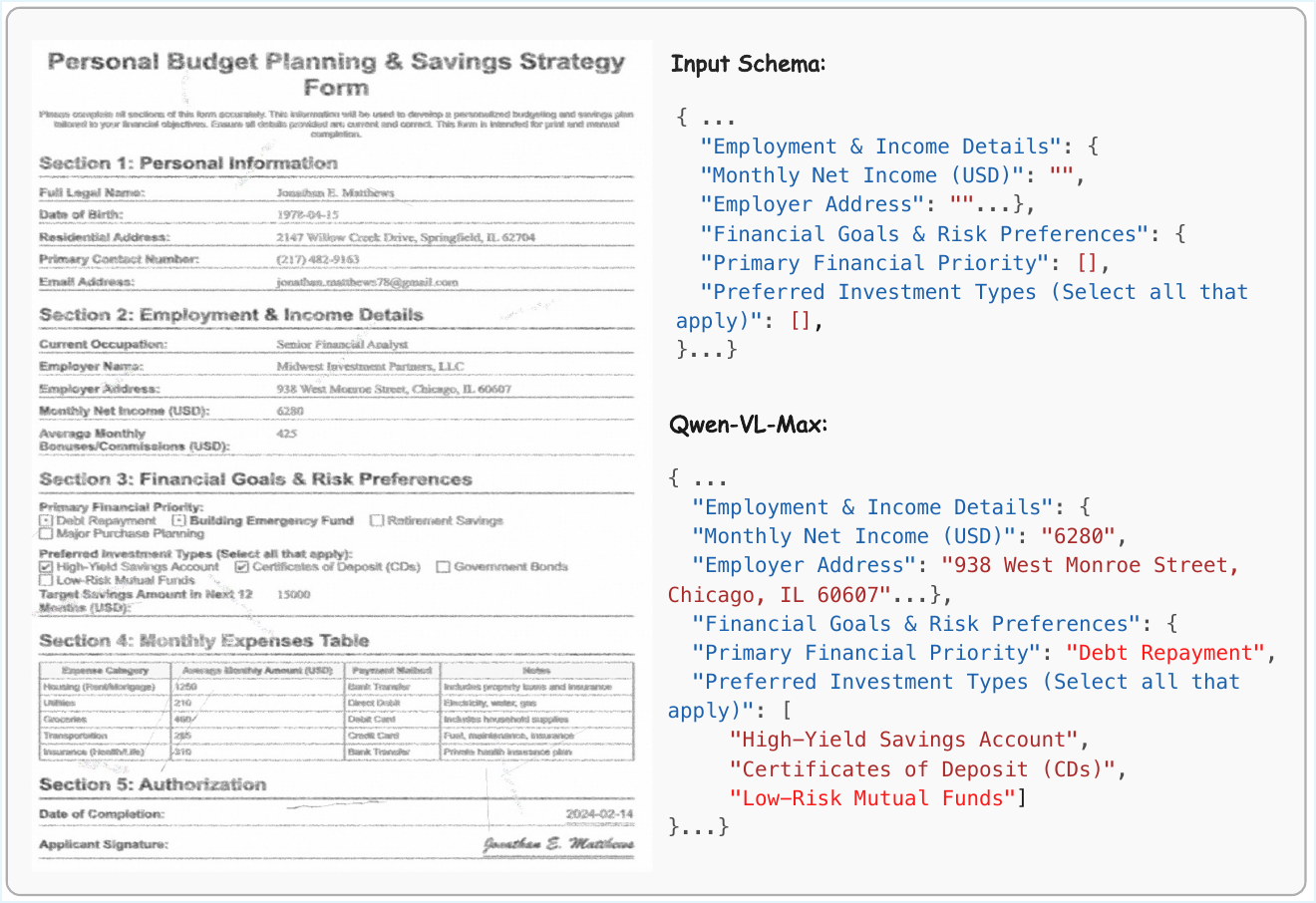}
    \caption{Example from the Form Document Type in the Open-Category KIE Track.}
    \label{fig:case2}
\end{figure*}

Figure~\ref{fig:case2} presents an example from the Form document type in the open-category KIE track. In contrast to receipts, this document exhibits a highly structured, multi-section layout, with fields distributed across well-defined regions, including free-text inputs, tabular expense records, and checkbox-based selections. 
This document type presents a distinct challenge for LMMs, as successful extraction requires not only accurate textual recognition but also precise layout-aware reasoning over heterogeneous field types. 
As illustrated by the example prediction, the model correctly recovers most fields, yet exhibits layout perception errors on selection-based fields, such as misidentifying or over-selecting checkbox options. 
These failure cases indicate that LMMs struggle to reliably associate visual selection cues with fields, underscoring the critical role of fine-grained layout understanding in KIE.

\begin{figure*}
    \centering
    \includegraphics[width=\linewidth]{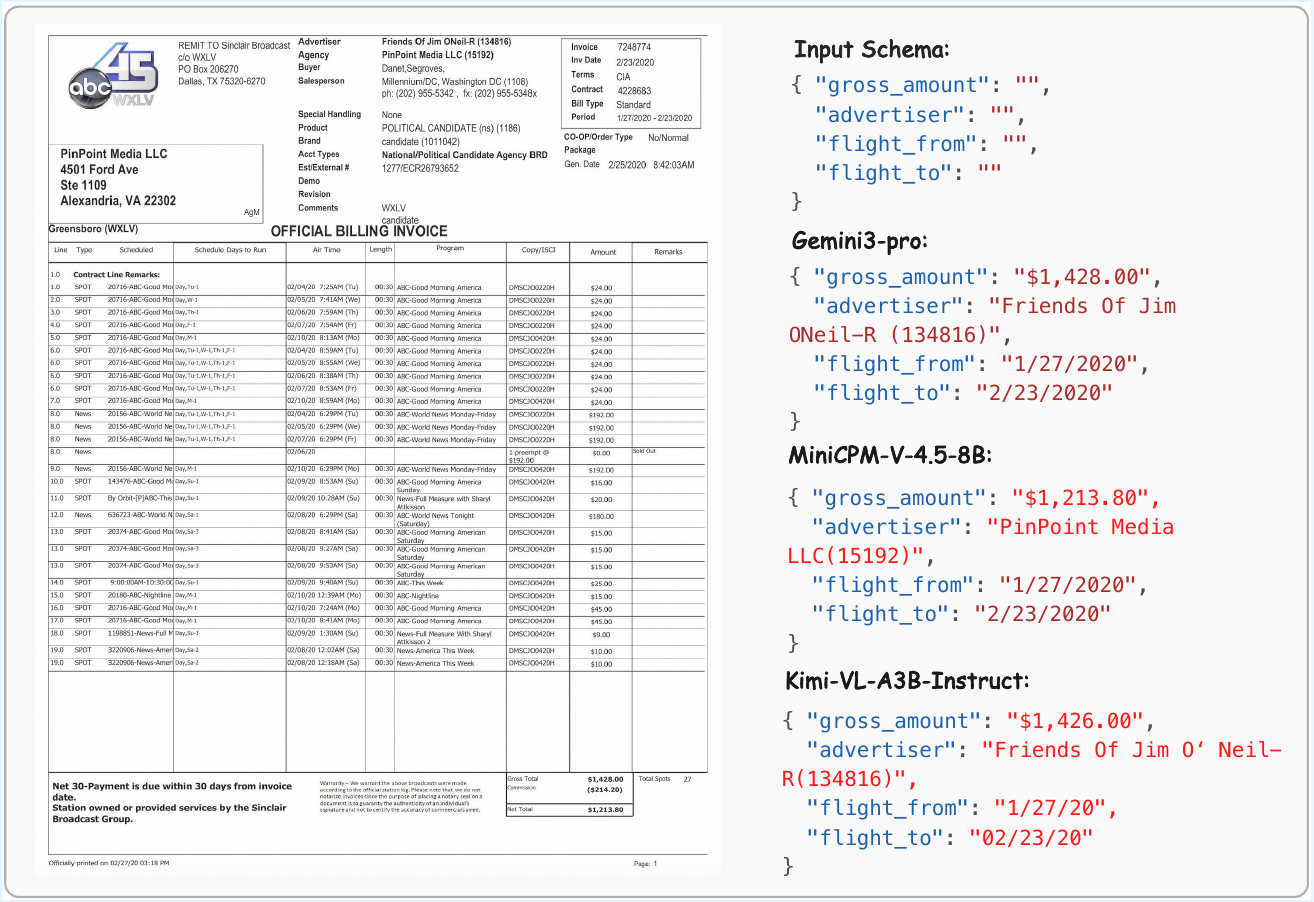}
    \caption{Example from the Advertisement Scenario in the Constrained-Category KIE Track.}
    \label{fig:case3}
\end{figure*}

Figure~\ref{fig:case3} illustrates an example from the constrained-category KIE track under the advertisement scenario. The document is visually dense and information-rich, integrating broadcast metadata, advertiser and agency information, detailed schedule tables, and summary totals within a single invoice layout. 
While the layout appears regular, key fields are distributed across disparate regions of the page and expressed in heterogeneous formats. 
LMMs exhibit inconsistent extraction behaviors: some correctly recover the advertiser and flight dates but misestimate the gross amount, while others confuse semantically related entities. These errors suggest that robust KIE remains challenging due to subtle layout dependencies, long-range field–value associations, and the need to disambiguate closely related entities in complex business documents.

\begin{figure*}
    \centering
    \includegraphics[width=\linewidth]{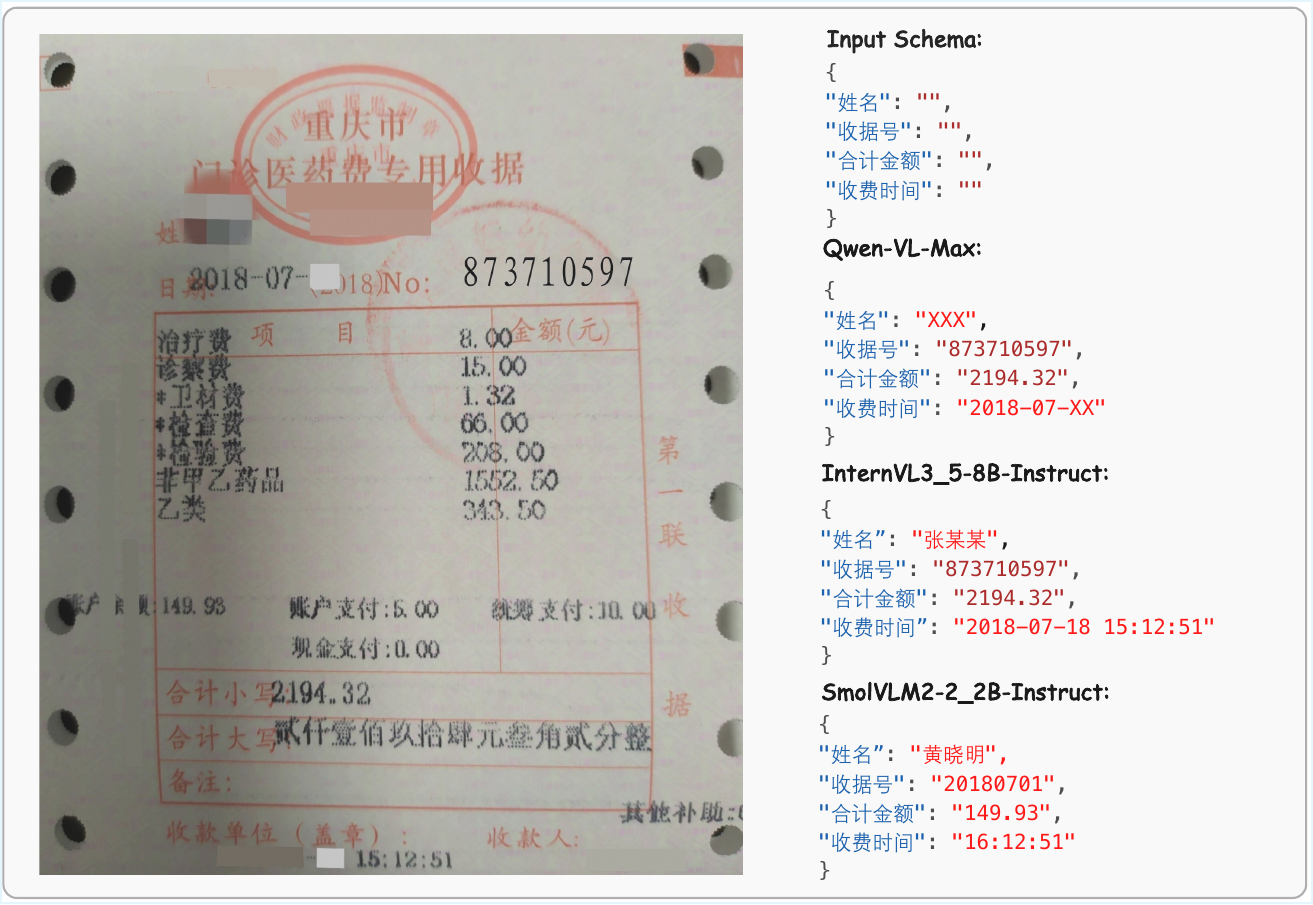}
    \caption{Example from the Medical Services Scenario in the Constrained-Category KIE Track.}
    \label{fig:case4}
\end{figure*}

Figure~\ref{fig:case4} shows a medical services example from the constrained-category KIE track. In practical applications, such documents are frequently shared, stored, or processed with sensitive regions (\textit{e.g.}, personal identifiers) manually redacted for privacy protection. In this example, the redaction removes critical textual evidence required by the input schema.
This commonly occurring masking practice can trigger severe hallucination behaviors in LMMs. When the ground-truth value is partially or fully occluded, models may still produce confident yet unsupported predictions, effectively inventing missing content rather than abstaining. As reflected in the predictions, some LMMs hallucinate plausible personal names that are not present in the visible document, while others substitute unrelated numeric fragments as the receipt number, total amount, or billing time.
This example highlights a practical and underexplored failure mode in real-world KIE deployments: even under a constrained schema, the widespread use of privacy-driven redaction can break visual grounding and lead to fluent but incorrect extractions.

\subsection{Instructions for Annotators}
We employ two distinct annotation workflows for the constrained-category and open-category KIE tracks. The annotation process was conducted by three collaborators of this work, who voluntarily participated as annotators. All annotators held master’s degrees, had 1–3 years of experience in real-world enterprise KIE systems, and were proficient in both Chinese and English. All annotations are completed on the itag\footnote{\url{https://www.aliyun.com/product/bigdata/learn/itag}} platform.

\begin{figure*}
    \centering
    \includegraphics[width=\linewidth]{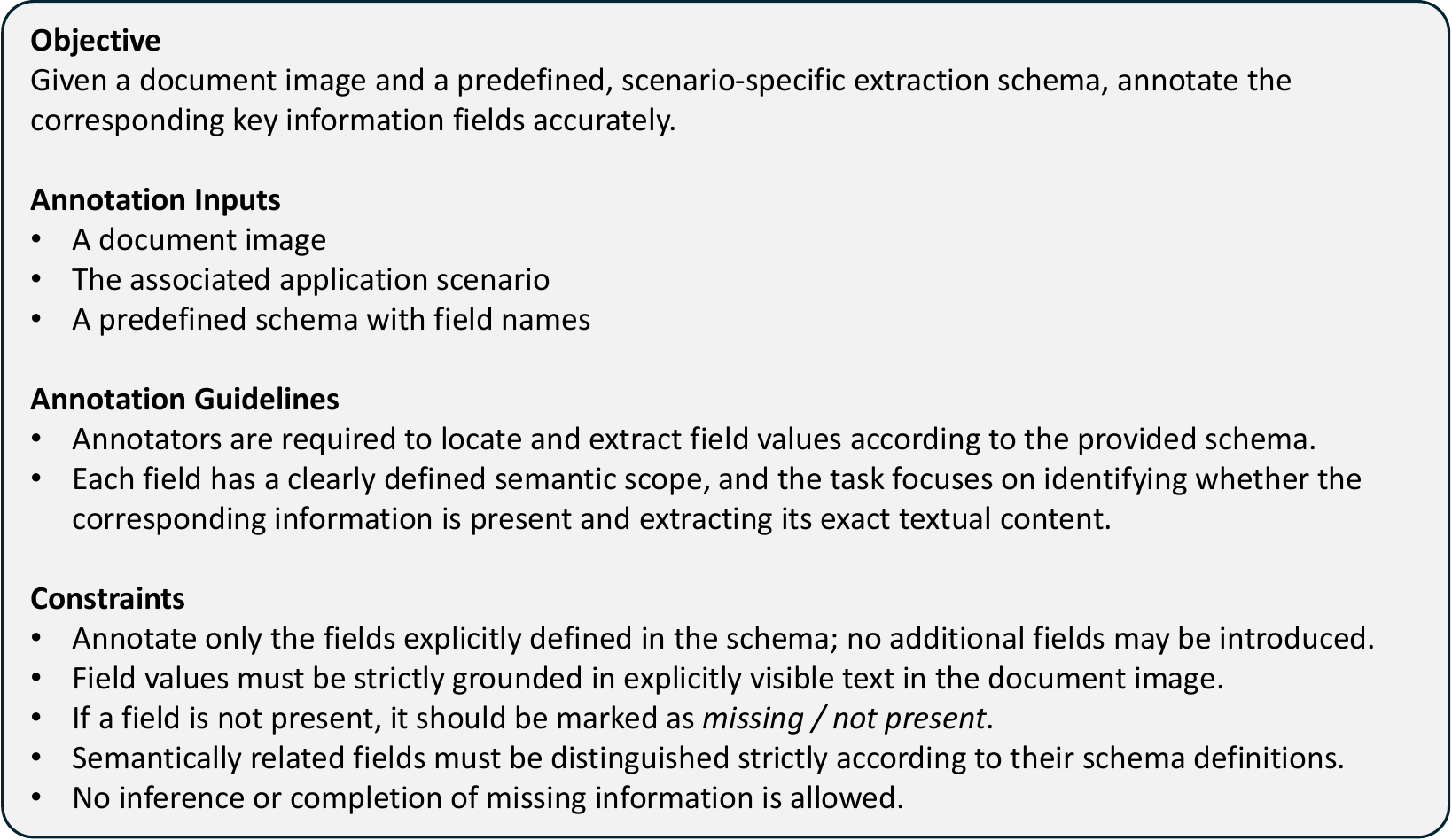}
    \caption{Instructions for Annotators in the Constrained-Category KIE Track.}
    \label{fig:ins_con}
\end{figure*}

The instruction for annotators in the constrained-category KIE track is shown in Figure~\ref{fig:ins_con}. Annotators follow a predefined schema and extract the fields defined therein, grounding each field value in visible document text and marking fields as missing when the corresponding information is absent.

\begin{figure*}
    \centering
    \includegraphics[width=\linewidth]{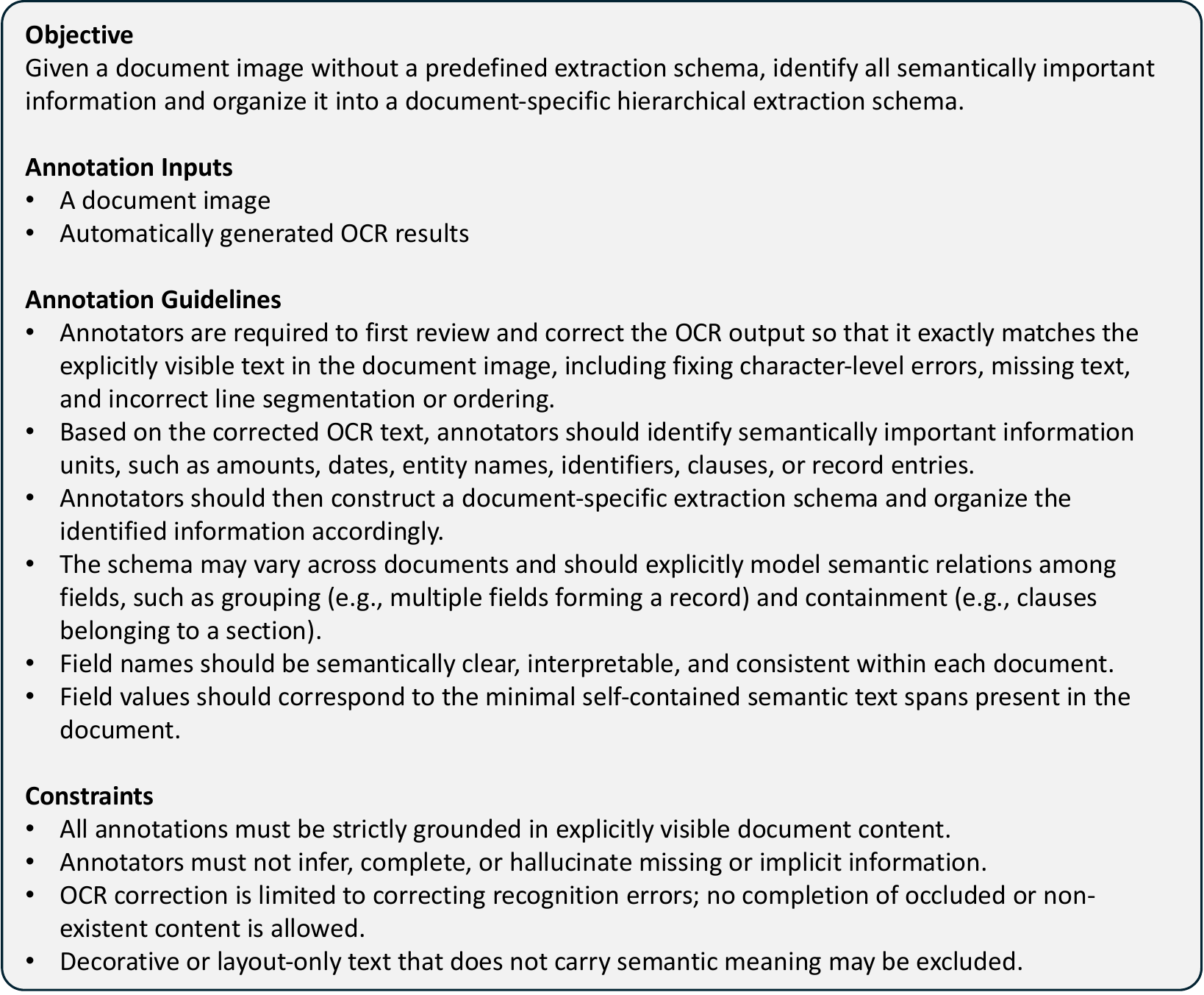}
    \caption{Instructions for Annotators in the Open-Category KIE Track.}
    \label{fig:ins_open}
\end{figure*}

For the open-category KIE track, no predefined extraction schema is provided. As shown in Figure~\ref{fig:ins_open}, annotators are required to review and correct the OCR output so that it exactly matches the visible document text, and then identify all key information units present in the document. Based on these units, annotators construct a document-specific hierarchical extraction schema that organizes fields and their semantic relations, such as grouping and containment. All annotations must remain strictly grounded in observable content, without inference, completion, or hallucination.

\end{document}